\documentclass{article}

\usepackage[preprint,nonatbib]{neurips_2024}

\usepackage[utf8]{inputenc} 
\usepackage[T1]{fontenc}    
\usepackage{hyperref}       
\usepackage{url}            
\usepackage{booktabs}       
\usepackage{amsfonts}       
\usepackage{nicefrac}       
\usepackage{microtype}      
\usepackage{xcolor}         
\usepackage{graphicx}         
\usepackage{amsmath}        
\usepackage{multirow}        
\usepackage{amssymb}        
\usepackage{algpseudocode}
\usepackage{algorithm}
\usepackage{wrapfig}

\title{AnyFit: Controllable Virtual Try-on for Any Combination of Attire Across Any Scenario}

\author{
\small 
Yuhan Li$^{1}$\thanks{work done during an internship at Alibaba.}, Hao Zhou$^{2}$, Wenxiang Shang$^{2}$, Ran Lin$^{2}$, Xuanhong Chen$^{1}$, Bingbing Ni$^{1}$\thanks{corresponding authors.}
\\[2pt] \small 
$^{1}$Shanghai Jiao Tong University, Shanghai 200240, China
\\[2pt] \small 
$^{2}$Alibaba 
\\[2pt] \small 
\{melodious, nibingbing\}@sjtu.edu.cn \\
\small 
\url{https://colorful-liyu.github.io/anyfit-page/} \\
}

\begin{document}

\maketitle

\begin{figure}[h!]
  \centering
  \vspace{-7mm}
  \includegraphics[width=\linewidth]{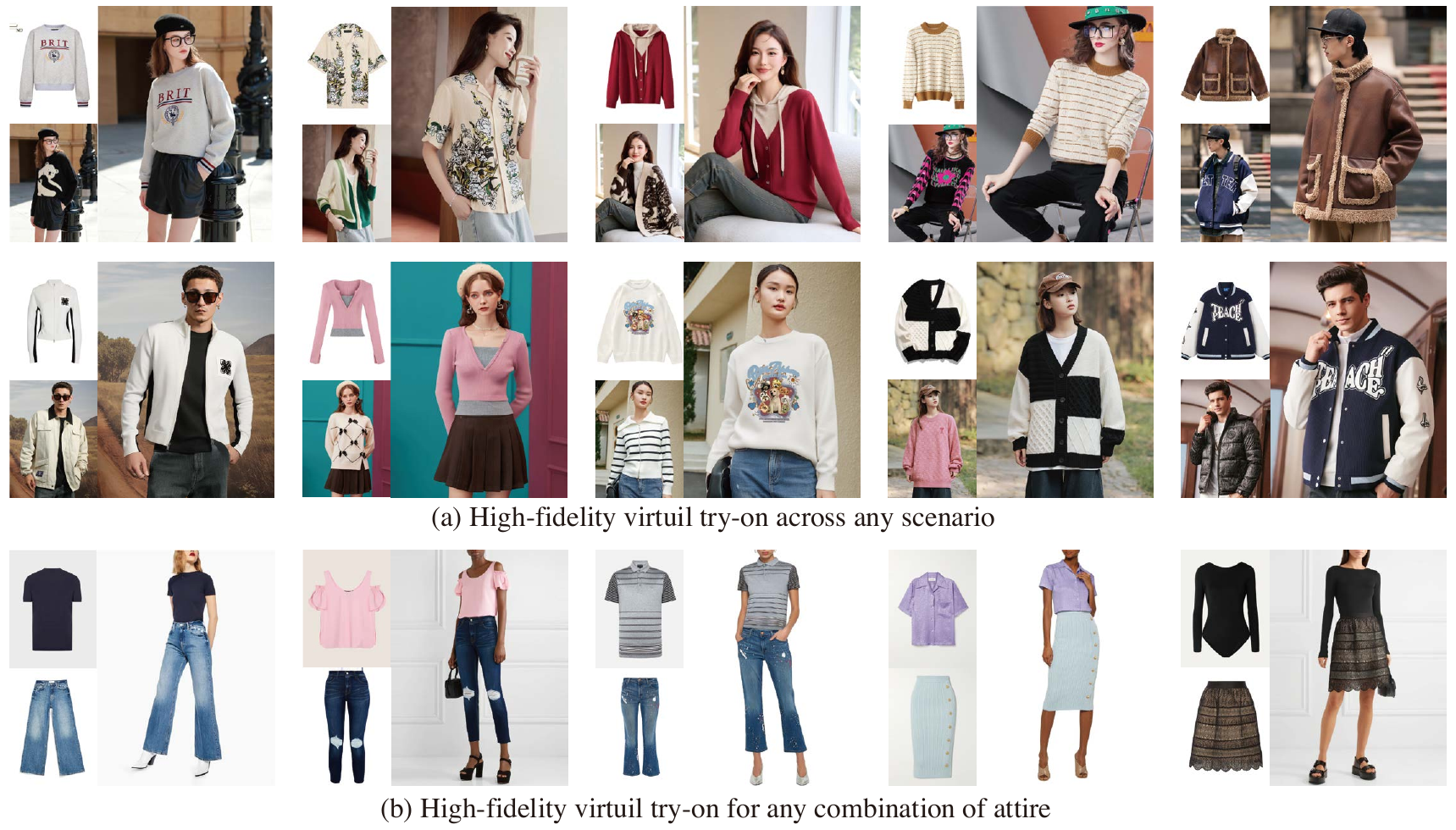}
  \vspace{-5mm}
  \caption{AnyFit shows superior try-ons for any combination of attire across any scenario.}
  \label{fig:showcase}
  
\end{figure}

\begin{abstract}
While image-based virtual try-on has made significant strides, emerging approaches still fall short of delivering high-fidelity and robust fitting images across various scenarios, as their models suffer from issues of ill-fitted garment styles and quality degrading during the training process, not to mention the lack of support for various combinations of attire. Therefore, we first propose a lightweight, scalable, operator known as Hydra Block for attire combinations. This is achieved through a parallel attention mechanism that facilitates the feature injection of multiple garments from conditionally encoded branches into the main network. Secondly, to significantly enhance the model's robustness and expressiveness in real-world scenarios, we evolve its potential across diverse settings by synthesizing the residuals of multiple models, as well as implementing a mask region boost strategy to overcome the instability caused by information leakage in existing models. 
Equipped with the above design, AnyFit surpasses all baselines on high-resolution benchmarks and real-world data by a large gap, excelling in producing well-fitting garments replete with photorealistic and rich details. Furthermore, AnyFit’s impressive performance on high-fidelity virtual try-ons in any scenario from any image, paves a new path for future research within the fashion community.
\end{abstract}

\section{Introduction}
\vspace{-2mm}
The dramatic success of e-commerce is steadily demanding a more convenient and personalized customer shopping experience. Among them, image-based virtual try-on (VTON)~\cite{han2018viton, wang2018toward, han2019clothflow} has emerged as a promising topic in the research community and witnesses rapid advancements~\cite{cui2021dressing, he2022fs_vton, Ge2021PFAFN}, whose task is to fit the target garment to the human body with various gestures. However, current methodologies do not meet the high-fidelity and robustness required in real-world applications, often resulting in artifacts or the mismatch of clothing details. Furthermore, the support for a variety of try-on combinations of attire~\cite{chen2024wearanyway} remains an area of ongoing research.

 Most prior methodologies~\cite{bai2022sdafn, chopra2021zflow} utilize a separate warping module to align garments on the human body, subsequently employing a Generative Adversarial Network (GAN)~\cite{goodfellow2014GAN} for their integration. This explicit warping process typically yields overly smooth garment transformations and struggles to cope with complex poses and occlusions.~\cite{xu2024ootdiffusion}. 
 While some diffusion-based methods~\cite{zhu2023tryondiffusion,xu2024ootdiffusion} leverage pre-trained diffusion models~\cite{rombach2022ldm}, with structures akin to ReferenceNet~\cite{hu2023animate} to preserve fine-grained garment information; However, these methods encounter difficulties in producing vivid fabric textures and photorealistic lighting and shadows. They also show artifacts in cross-category try-ons, diminishing their inherent text-image capacity when applied to specialized tasks as depicted in Fig.~\ref{fig:main_wild}.
 In summary, existing methods still fall short in producing images of high fidelity that exhibit clothing styles rendered with exceptional detail and true-to-life accuracy across scenes. Moreover, these methods are designed solely for trying on individual items of clothing and do not support multi-conditions, thereby failing to facilitate the free combination of tops and bottoms. 

As discussed above, we believe that an ideal VTON workflow should exhibit the following properties:
\begin{itemize}
\vspace{-2mm}
\item \textbf{Scalability.} The ultimate goal of the VTON model is to enable any free and customizable virtual outfit combination of multiple garments~\cite{zhang2024mmtryon}. It should support multi-condition injection, allowing for easy expansion to more applications, such as mixing and matching tops and bottoms, layering inner and outer garments, \emph{etc}.
\item \textbf{Robustness.} Given the diverse scenarios encountered in e-commerce settings~\cite{choi2024idm}, the VTON model should generate authentic fabric textures and natural lighting, reproducing the details of the target clothing (\emph{e.g.}, logos, patterns, texts and strips) stably and accurately.
\end{itemize}
\vspace{-2mm}

We present the following critical contributions to establish AnyFit as a novel VTON paradigm, which adeptly addresses the challenge of any combination of attire across any conceivable scenario, in Fig.~\ref{fig:showcase}. AnyFit mainly consists of two isomorphic U-Nets, namely HydraNet and MainNet. The former is tasked with extracting fine-grained clothing features, while the latter is responsible for generating try-ons. (1) \textbf{Scalability}: A hallmark of AnyFit is its innovative introduction of the \textbf{Hydra Encoding Block} that only parallelizes attention matrices within a sharing HydraNet, enabling effortless expansion to any quantity of conditions with only 8\% increase in parameters for each additional branch. The proposal of parallelizing these blocks is built on the insight that only the self-attention layers are crucial for implicit warping~\cite{zhu2023tryondiffusion}, while the remaining components primarily serve as generic feature extractors. We further invent \textbf{Hydra Fusion Block} to seamlessly integrate the features of Hydra Encoding into MainNet, with positional embeddings to distinguish encodings from different sources. It is important to note that ReferenceNet~\cite{hu2023animate, chang2023magicdance} or GarmentNet~\cite{xu2024ootdiffusion} could be seen as specific instances of HydraNet when limited to a single condition.
(2) \textbf{Robustness}: Observations indicate a noticeable reduction in the robustness and quality of images generated by existing Virtual Try-On (VTON) works, in comparison to the original stable diffusion performances. Inspired by discussions in the community~\cite{ModelMerge}, we present the \textbf{Prior Model Evolution} strategy. This innovative approach involves merging parameter variations within a model family (\emph{e.g.}, a collection of fine-tuned versions of SDXL~\cite{podell2023sdxl}), enabling the independent evolution of multiple capabilities of the base model.
This strategy emerges as an intuitively logical and highly effective method for amplifying the model's innate potential prior to training. This is particularly relevant when contending with the significant escalation in training costs associated with dual U-Nets—an aspect that is overlooked in previous research.
Furthermore, we introduce the \textbf{Adaptive Mask Boost} to further enhance the fit of the attire as a bonus. It requires length augmentation of \emph{parsing-free} mask regions during the training phase, allowing the model to autonomously understand the overall shape of the clothing, which emancipates the model from previous reliance on hints of masks derived from garments. During inference, we adapt the shape of the mask area based on the aspect ratio of the target garment, thereby markedly encouraging the generation of well-fitted try-ons, particularly for long garments (\emph{e.g.}, windbreaker).

To the best of our knowledge, AnyFit stands as a pioneering VTON method to fulfill scalability and robustness requirements. 
Our innovative HydraNet and Prior Model Evolution strategies have the potential to transform not just the domain of VTON, but to catalyze advancements across a broader spectrum of conditional generation applications.
Finally, we have carried out comprehensive experiments on try-on benchmarks~\cite{choi2021vitonhd, morelli2022DressCode} and engaged in challenging validation using in-the-wild sets. These experiments demonstrate that our model shows exceptional performance that eclipses current methods by a substantial margin, in terms of garment fidelity and robustness when addressing street-captured scenarios. In addition, our method has realized a formidable capability for multi-garment try-ons, culminating in results that exhibit strikingly harmonized upper and lower styles.

\section{Related works}

\noindent \textbf{GAN-based virtual try-on.} The virtual try-on task is concerned with synthesizing images of a person donning the designated garment with appropriate fit~\cite{han2018viton}, while retaining salient characteristics of the original garment and person, given a pair of images depicting a person and a target garment. To execute this task, numerous works~\cite{dong2020fashion, jo2019scfegan, lee2020maskgan, bai2022sdafn, chopra2021zflow, han2019clothflow} have utilized Generative Adversarial Networks (GANs)~\cite{goodfellow2014GAN} with two-stage strategy~\cite{lee2022hrviton, Ge2021PFAFN, yang2020acgpn}: (1) warping the clothing to the desired shape~\cite{bookstein1989tps, lin2017fpn} and (2) fusing the deformed clothing via try-on generator based on GAN. HR-VITON~\cite{lee2022hrviton} conducts both warping and segmentation concurrently to address issues related to body occlusion and misfit of garments. GP-VTON~\cite{xie2023gpvton} introduces localized warping along with global parsing to independently simulate the deformation of different regions of clothing, aiming to achieve a more form-fitting result. However, these existing approaches that rely on an explicit warping module are incapable of supplementing the sides of the clothing and the natural lighting and shadows~\cite{xu2024ootdiffusion}. 

\noindent \textbf{Diffusion-based virtual try-on.} As significant progress in Text-to-Image diffusion models~\cite{ho2020ddpm, nichol2021glide, huang2023composer, chen2023anydoor} is witnessed in recent years, some works~\cite{chen2024wearanyway} have been motivated to incorporate pre-trained diffusion models~\cite{rombach2022ldm,podell2023sdxl} as priors into virtual try-on task. LADI-VTON~\cite{morelli2023ladi} and DCI-VTON~\cite{gou2023dci} explicitly deform the clothing to achieve pixel-level alignment with the human body, followed by a diffusion model to blend the clothing with the human body as refinement. StableVITON~\cite{kim2023stableviton} introduces an end-to-end approach that injects intermediate feature maps from a spatial encoder into the U-Net decoder via a zero cross-attention block, akin to the ControlNet~\cite{zhang2023controlnet} structure. Most recently, OOTDiffusion~\cite{xu2024ootdiffusion} and IDM~\cite{choi2024idm} achieve garment feature extraction with a parallel U-Net and feed them through self-attention for enhanced integration. Unfortunately, these methods intrinsically lack support for try-ons that involve multiple garments. Moreover, they exhibit artifacts and unstable garment fits for arbitrary images, which leads to a degradation of performance on out-of-distribution images in complex backgrounds and poses.

\vspace{-2mm}
\section{Method}
\vspace{-2mm}
\begin{figure}[t]
  \centering
  \vspace{-7mm}
  \includegraphics[width=\linewidth]{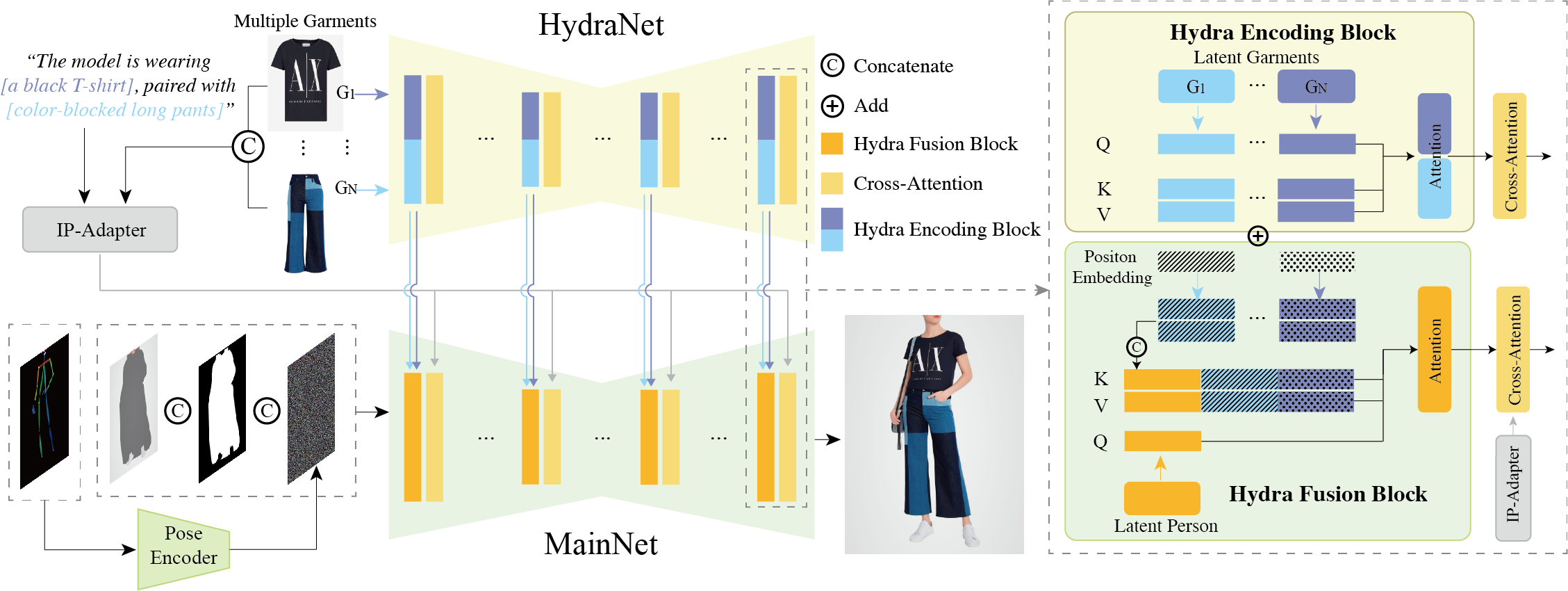}
  \vspace{-5mm}
  \caption{Overall framework of our method.}
  \vspace{-5mm}
  \label{fig:framework}
\end{figure}

\subsection{Model overview}
\vspace{-2mm}

An overview of the AnyFit is presented in Fig.~\ref{fig:framework}. The backbone of AnyFit employs the SDXL~\cite{rombach2022ldm}, with the preliminary detailed in Appendix~\ref{appendix:preliminary}.
Given a human image $x_h \in \mathbb{R}^{H\times W\times 3}$ and a target garment image $x_g \in \mathbb{R}^{H \times W\times 3}$, AnyFit is aimed to generate an authentic try-on image $x_{tr}$. We employ OpenPose~\cite{cao2017openpose, yang2023effective} to obtain clothing-agnostic mask $x_m$ and masked person image $x_{ag}$ adjusting for the size of different garments, as detailed in Sec.~\ref{sec:robust}. We treat VTON as a specific case of image inpainting~\cite{yang2023paintbyexample}, endeavoring to fill the masked person $x_{ag}$ with the cloth $x_{g}$. The main inpainting U-Net (MainNet) inputs $3$ concatenated components with $9$ channels: the noisy image $z_t$, the latent agnostic image $E(x_{ag})$ and the resized agnostic mask $x_m$, where $E(\cdot)$ represents VAE~\cite{kingma2013vae} encoding. A Pose Guider~\cite{hu2023animate} with $4$ convolution layers ($4\times 4$ kernels, $2\times 2$ strides,  $16,32,64,128$ channels) is incorporated to align the pose image $E(x_p)$ with noise $z_t$.

\textbf{Scalability:} To preserve the fine details of the clothing, as well as to support both single and multiple garment VTONs, we employ a HydraNet that mirrors the MainNet in encoding clothing information. It shares the same weight initialization as the MainNet and innovatively parallelizes attention metrics based on the number of conditions to create Hydra Encoding Blocks for different conditional encodings. \textbf{Robustness:} During training, issues such as mask information leakage and quality degradation were observed. To address these issues, we adopt Adaptive Mask Boost and Prior Model Evolution, respectively, which significantly bolster the model's robustness across different scenarios cost-effectively and straightforwardly.

\vspace{-2mm}
\subsection{HydraNet for multi-condition VTON}
\vspace{-2mm}

\noindent \textbf{HydraNet.} 
Inspired by successful practice in human editing~\cite{hu2023animate, chang2023magicdance}, we introduce a garment encoding network isomorphic to the main generative network (MainNet), that precisely preserves the details of clothing. When dealing with multi-garment VTON, a direct method might involve replicating multiple garment encoding nets to manage different conditions. This approach, however, would lead to a significant increase in the number of parameters, rendering it computationally prohibitive. Experimentally we discover that for conditions with similar content (such as different types of clothing), the self-attention module plays a vital role in the latent warping of the garments, aligning them with the locations requiring inpainting. Conversely, other network architectures, which typically tasked with general feature extraction, can be shared across different condition encoding branches without compromising the model's performance.
In view of this, we innovatively propose HydraNet for multi-condition encoding. It operates based on a shared Unet structure, while parallelizing the attention modules according to the number of input conditions, thereby constructing Hydra Encoding Blocks.
Specifically, we parallelize the self-attention matrices with identical initial weights, and feed the multi-condition \emph{key} and \emph{value} features $\left \{ z_{hk}^i, z_{hv}^i \right \}$ into MainNet, which encode the fine-grained details of the clothing. It's notable that ReferenceNet~\cite{hu2023animate, chang2023magicdance} or GarmentNet~\cite{xu2024ootdiffusion} could be seen as specific instances of HydraNet limited to a single condition. HydraNet requires only one forward pass (timestep $t=0$) to encode clothing before the multiple denoising steps in MainNet, with the additional temporal and parameter overheads being minimal for each added condition, in Tab.~\ref{tab:multi}.

\noindent \textbf{Hydra Fusion.}
We propose a highly efficient and easily scalable Hydra Fusion Block to replace the self-attention layers in MainNet, accomplishing the feature injection from HydraNet to MainNet for any length via concatenation. Specifically, given the \emph{key} and \emph{value} features $\left \{ z_{hk}^i, z_{hv}^i \right \} \in \mathbb{R}^{b\times l\times c}$ from the HydraNet, we introduce learnable position embeddings to distinguish features from different source conditions. The superscript $i$ denotes different input conditions. Subsequently, we concatenate the \emph{key} and \emph{value} along the $l$ dimension to obtain the final $\left \{ z_{hk}^{all}, z_{hv}^{all} \right \} \in \mathbb{R}^{b\times N l\times c}$ as:
\begin{equation}
z_{hk}^{all} = ( z_{hk}^1 + PE^1(z_{hk}^1) ) \oplus (z_{hk}^2 + PE^2(z_{hk}^2))\oplus \dots \oplus (z_{hk}^N + PE^N(z_{hk}^N)),
\end{equation}
where $N$ denotes the total number of input conditions, $PE$ represents positional encoding, and $\oplus$ signifies concatenation. $z_{hv}^{all}$ follows a similar formulation. Facing the \emph{key} and \emph{value} features $\left \{ z_{mk}, z_{mv} \right \} \in \mathbb{R}^{b\times l \times c}$ from the MainNet and concatenated HydraNet features $\left \{ z_{hk}^{all}, z_{hv}^{all} \right \}$, we once again concatenate the corresponding features along the $l$ dimension into $\left \{ z_{ck}, z_{cv} \right \} \in \mathbb{R}^{b\times \left (N+1 \right ) l\times c}$, which are then used in subsequent attention calculations with $z_{mq}$.
It is noteworthy that, with the lightweight design leveraging parallelization and concatenation, HydraNet can be effortlessly extended to perform injections with any number of conditions, thereby possessing a more expansive application potential within the domain of generative models.

\begin{figure}[ht]
  \centering
\includegraphics[width=1.0\textwidth]{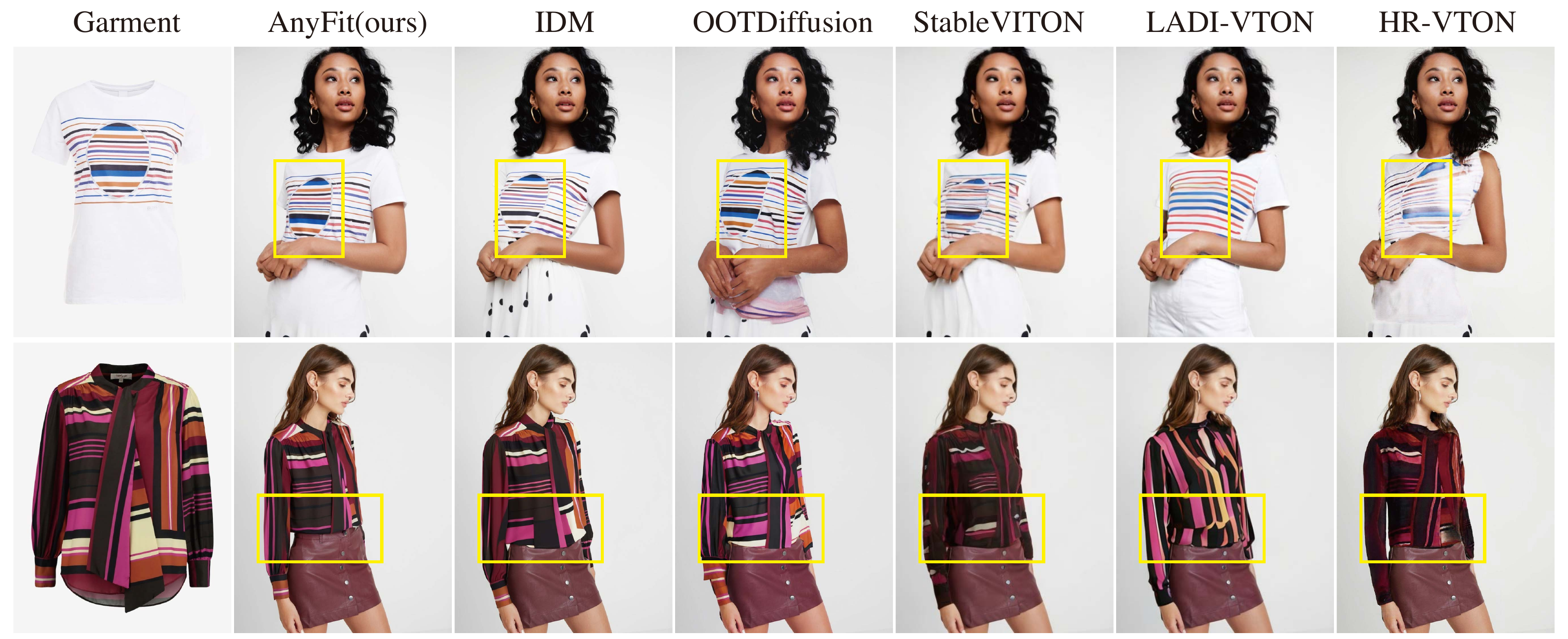}
\vspace{-5mm}
  \caption{Visual comparisons on VITON-HD. AnyFit displays superior details and outfit styling.} 
\vspace{-5mm}
  \label{fig:main_viton}
\end{figure}

\vspace{-2mm}
\subsection{Model evolution and mask boost for robust VTON}
\label{sec:robust}
\vspace{-2mm}

\noindent \textbf{Prior Model Evolution.}
A diminution in image generation performance with the SDXL-inpainting model compared to the SDXL base model is noted~\cite{worse_inpainting}. We attribute this degradation to the disruption of previously well-aligned correspondences between text and images during the inpainting pre-training phase. Drawing inspiration from the open-source community~\cite{ModelMerge}, we develop a Prior Model Evolution strategy, which enhances the model's strength and adaptability in generating outfit images at a very low cost, even without training. Specifically, we meticulously amalgamate the weights from three distinct, powerful models to evolve the initial weights for our model. These models comprise: the SDXL-base-1.0~\cite{podell2023sdxl}, the SDXL-inpainting-0.1~\cite{sdxl_inp} with inpainting capabilities, and DreamshaperXL alpha2~\cite{dreamshaperxl}, which demonstrates superior performance in generating clothing and human figures. The parameter weights of these models are denoted by $\mathbf{W}_{base}, \mathbf{W}_{inp}, \mathbf{W}_{ds}$ respectively. The formula for the evolution is as follows:  
\vspace{-2mm}
\begin{equation}
\mathbf{W}_{new} = \mathbf{W}_{base} + \alpha \cdot \underset{\mathrm{Inpainting \, Increment}}{\underbrace{\left ( \mathbf{W}_{inp} - \mathbf{W}_{base} \right )}}  + \beta \cdot \underset{\mathrm{Outfitting \, Increment}} {\underbrace{\left ( \mathbf{W}_{ds} - \mathbf{W}_{base} \right )}}  ,
\label{eq:merge}
\end{equation}
where $\alpha$ and $\beta$ are the balancing coefficients that account for the capability enhancements from SDXL-inpainting and DreamshaperXL. It is important to note that we directly copy the extra $5$ channels in the \emph{conv in} layer of the SDXL-inpainting into the merged model, multiplying them by $\alpha$.

However, the optimal values of $\alpha$ and $\beta$ are not apparent. We hope to determine the optimal $\alpha$ and $\beta$ to ensure that the initial weight $\mathbf{W}_{new}$ achieve the best evaluation performance, i.e. 
\vspace{-2mm}
\begin{equation}
\mathop{\arg\min}_{(\alpha, \beta)\in [0,2]^2} f(\alpha, \beta) =  \Phi\left( \mathbf{W}_{base} + 
\alpha \cdot \left ( \mathbf{W}_{inp} - \mathbf{W}_{base} \right )
+ \beta \cdot  \left ( \mathbf{W}_{ds} - \mathbf{W}_{base} \right) \right),
\end{equation}
where $\Phi$ is a non-differentiable evaluation function. Empirically, we assume that $f$ exhibits monotonic or convex properties with respect to the balancing coefficients $(\alpha, \beta)$ in most regions. 
Therefore, we discretize the continuous domain $[0,2]^2$ into a grid with $\delta=0.1$ as the step size and design the discrete greedy algorithm 1, to search for the optimal $(\alpha, \beta)$. In our algorithm, we selecte the CLIP score~\cite{radford2021clip} on $20$ fixed inpainting image-text pairs as the evaluation function $\Phi$. The optimal solution obtained is $(\alpha, \beta) = (1.0, 1.1)$. Please refer to the Appendix~\ref{appendix:algorithm} for more explanations.

\begin{wrapfigure}{l}{0.52\textwidth} 
\begin{minipage}{0.52\textwidth} 
\vspace{-8mm}
\begin{algorithm}[H]
\caption{Discrete greedy algorithm}
\begin{algorithmic}[1]
\fontsize{8pt}{9pt}\selectfont
\Require Evaluation function $f$, step size $\delta$.
\State Initialize $(\alpha, \beta) \gets (0.5, 0.5)$
\While{True}
    \State $f_{\text{current}} \gets f(\alpha, \beta)$
    \State $N \gets \{(\alpha + \delta, \beta), (\alpha - \delta, \beta), (\alpha, \beta + \delta), (\alpha, \beta - \delta)\}$ 
    \State $N \gets \{(\alpha', \beta') \in N \mid 0 \leq \alpha' \leq 2 \text{ and } 0 \leq \beta' \leq 2\}$ 
    \State $F_N \gets \{f(\alpha', \beta') \mid (\alpha', \beta') \in N\}$ 
    \If{$\min(F_N) \geq f_{\text{current}}$}
        \State \textbf{break}
    \Else
        \State $(\alpha, \beta) \gets \arg\min_{(\alpha', \beta') \in N} f(\alpha', \beta')$
    \EndIf
\EndWhile
\State \Return $(\alpha, \beta)$
\end{algorithmic}
\end{algorithm}
\label{algrithm}
\vspace{-8mm}
\end{minipage}
\end{wrapfigure}
\begin{figure}[t]
  \centering
\includegraphics[width=1.0\textwidth]{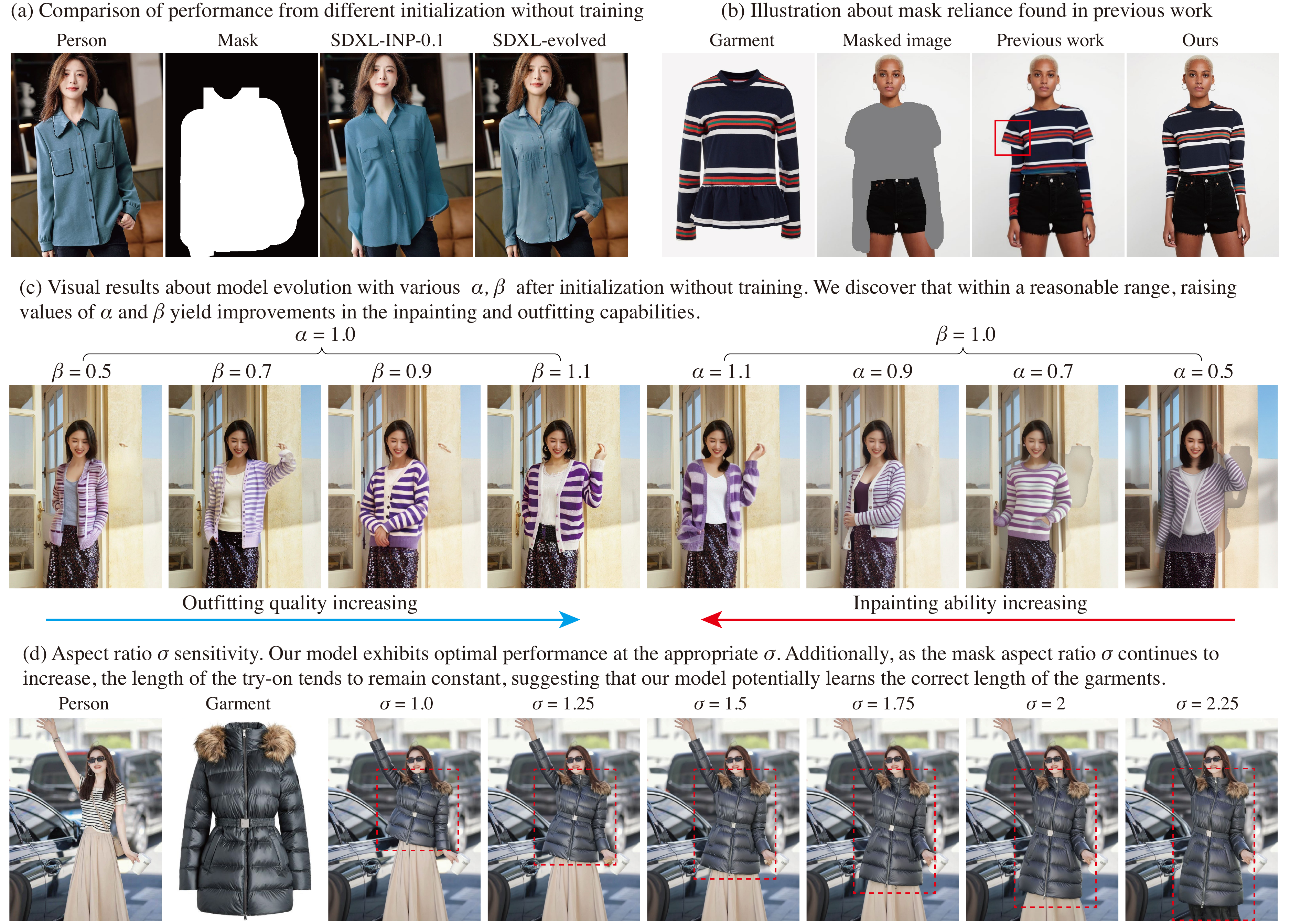}
  \caption{Visual validation about model evolution and mask boost in (a), (c), (d). We also provide visual results about mask reliance in (b) found in previous work.
} 
  \label{fig:adaptive_mask}
\end{figure}

\noindent \textbf{Adaptive Mask Boost.}
Previous works generally exhibit limited robustness in cross-category try-on scenarios, resulting in inaccurately rendered clothing styles as shown in Fig.~\ref{fig:adaptive_mask} and ~\ref{fig:parsing_leaky}. This is largely due to a dependence on agnostic masks derived from clothing parsing, which tends to leak the edges of the clothing shape during training. This leakage may cause the generated garments to almost entirely cover the agnostic mask region. In response to these limitations, we employ an intuitive and effective approach that significantly enhances the model's robustness about cross-category try-on, \emph{i.e.}, the Adaptive Mask Boost strategy, which primarily comprises mask augmentation during training and adaptive elongation during inference. 
Specifically, during training, the agnostic mask is extracted solely using OpenPose body joint detections \emph{without leveraging human parsing}. We perform random elongation of the mask by a factor $f \sim \text{Uniform}(1.2, 1.5)$ with a probability of $P = 0.5$. This training setting forces the model to autonomously determine the optimal cloth length. During inference, we assess the aspect ratio $\sigma$ of the bounding box of the laid-out garment. If $\sigma > 1.2$, we proportionally extend the agnostic area to match $\sigma$, creating an adaptive agnostic mask that conforms to the garment's style. Experiments have validated that AnyFit with Adaptive Mask Boost autonomously determines the appropriate garment length, yielding robust try-on results across different clothing categories.

\begin{figure}[ht]
  \centering
\includegraphics[width=1.0\textwidth]{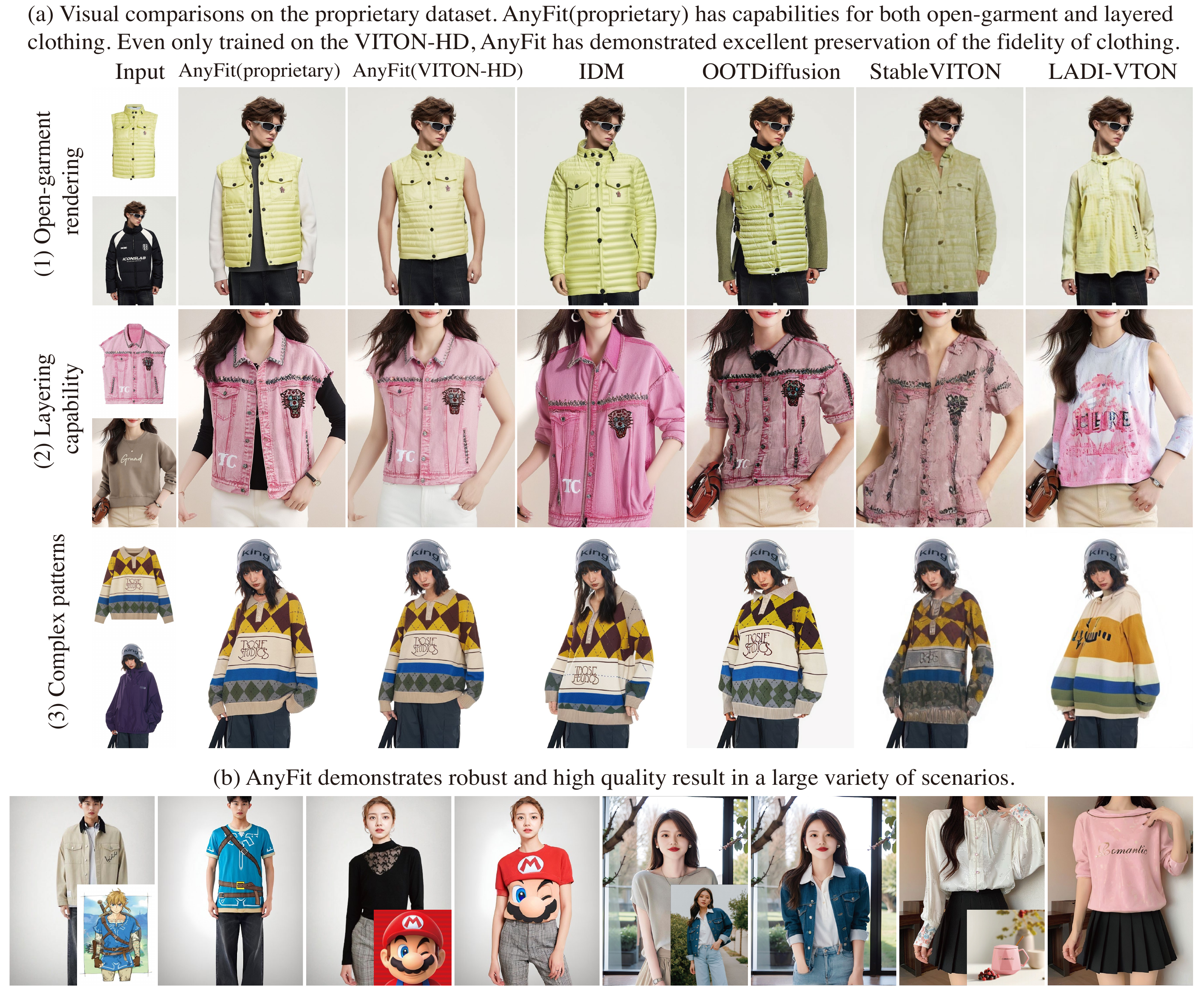}
  \caption{Visual results on proprietary and in-the-wild data. Best viewed when zoomed in.} 
  \label{fig:main_wild}
\end{figure}

\begin{figure}[t]
  \centering
\includegraphics[width=1.0\textwidth]{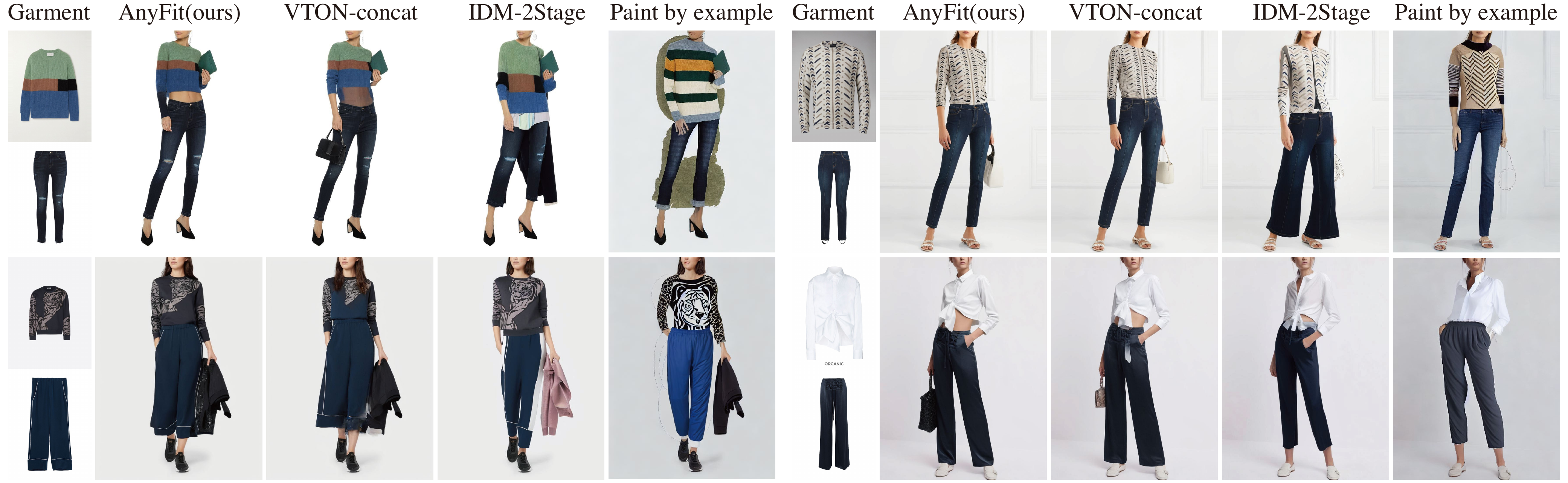}
  \caption{Visual comparisons on the DressCode-multiple. AnyFit exhibits an elegant integration between upper and lower garments, accurate length control, and appropriate overall styling.} 
  \label{fig:multi}
\end{figure}

\section{Experiments}

\subsection{Experimental setup}

\noindent \textbf{Datasets.} 
Our experiments are carried out on two publicly available datasets, VITON-HD~\cite{choi2021vitonhd} ($11647$ training pairs) and DressCode~\cite{morelli2022DressCode} ($48392$ training pairs) using the official splits for training and testing, as well as an additional proprietary e-commerce dataset. This proprietary dataset contains $50602$ training pairs and $2500$ testing pairs of mainly upper-body person and upper garment images, featuring complex patterns, backgrounds, and postures, alongside a rich variety of styles including layered garment ensembles, which present a more challenging scenario. As for multi-garment try-on, we utilized a HumanParsing model~\cite{jin2023sssegmenation, jin2024idrnet} to extract clothing items from DressCode and constructed triplets consisting of (upper-body garment, lower-body garment, model image). In these triplets, one garment is an original laid-out image, while the other is a cropped image from the person's image. Finally, we construct $24314$ publicly available upper-lower triplets by garment crop from the $15363$ upper-body pairs and $8951$ lower-body pairs in DressCode, called DressCode-multiple. A subset of $1800$ triplets is reserved as the test set. Please refer to the Appendix~\ref{appendix:datasets} for more details.

\noindent \textbf{Implementation details.}
We initialize the AnyFit with Prior Model Evolution strategy described in Sec.~\ref{sec:robust}, and fine-tune it using an AdamW optimizer~\cite{loshchilov2018adamw} with a constant learning rate of $5e-5$. We train three variants of models on the VITON-HD, DressCode and proprietary dataset at a resolution of $1024 \times 768$, independently. Subsequently, we extend the model trained on DressCode to enable multi-garment try-ons by training on the DressCode-multiple dataset. All the models are trained for $150$ epochs on $8$ NVIDIA A100 GPUs with DeepSpeed~\cite{deepspeed} ZeRO-2 to reduce memory usage, at a batch size of $15$. At inference time, we run AnyFit on a single NVIDIA RTX 3090 GPU for $30$ sampling steps with the DDIM sampler~\cite{song2020ddim}. Outfitting dropout~\cite{xu2024ootdiffusion} is used with a guidance scale $s_g = 1.3$. The data augmentation follows the same protocol as in StableVTON~\cite{kim2023stableviton}. We also employ pretrained IP-Adapter~\cite{ye2023ipadapter} for SDXL. Please refer to the Appendix~\ref{appendix:exp_details} for more details.

\noindent \textbf{Evaluation protocols.}
We measure reconstruction accuracy by LPIPS~\cite{zhang2018perceptual} and SSIM~\cite{wang2004ssim} in a paired setting provided ground truth images, and authenticity of unpaired synthesized images by FID~\cite{parmar2021cleanfid} and KID~\cite{sutherland2018kid} without ground truth. All evaluations are conducted at a resolution of $512 \times 384$.

\noindent \textbf{Baselines.}
We compare our model on single try-on tasks on VITON-HD, DressCode and our proprietary dataset with previous baselines including HR-VTON~\cite{lee2022hrviton}, LADI-VTON~\cite{morelli2023ladi}, DCI-VTON~\cite{gou2023dci}, StableVITON~\cite{kim2023stableviton}, OOTDiffusion~\cite{xu2024ootdiffusion}, and the state of the art IDM~\cite{choi2024idm}. We directly utilize their released pre-trained models.
As for multi-garment try-on, we developed a two-stage IDM model as a strong baseline, referred to as IDM-2Stage, which dresses the upper and lower garments sequentially. Inspired by \cite{chen2024wearanyway}, we concatenate the upper and lower garments on width spatially and feed them to a single-conditional HydraNet for training as another baseline, termed VTON-concat. Finally, we compared AnyFit with Paint by Example~\cite{yang2023paintbyexample}, IDM-2Stage, and VTON-concat.

\begin{table*}[t]
\caption{Quantitative comparisons on the VITON-HD~\cite{choi2021vitonhd} and DressCode~\cite{morelli2022DressCode}.}
\fontsize{8pt}{9pt}\selectfont
\begin{tabular}{c c c c c c c c c c c }
  \toprule
   \multicolumn{2}{c}{Dataset} & &\multicolumn{4}{c}{VITON-HD~\cite{choi2021vitonhd}} &\multicolumn{4}{c}{DressCode~\cite{morelli2022DressCode}}  \\
    \cmidrule{1-2} \cmidrule{4-11}  
  \multicolumn{2}{c}{Method}                              
  & &LPIPS $\downarrow$ & SSIM $\uparrow$ & FID $\downarrow$ & KID $\downarrow$&LPIPS $\downarrow$ & SSIM $\uparrow$ & FID $\downarrow$ & KID $\downarrow$  \\
    \cmidrule{1-2} \cmidrule{4-11} 
 \multicolumn{2}{c}{HR-VTON~\cite{lee2022hrviton}} 
  && 0.097 & 0.878 & 12.31 & 3.86 & - & - & - & - \\
  \multicolumn{2}{c}{DCI-VTON~\cite{gou2023dci}} 
  && \textbf{0.072} & \underline{0.892} & \underline{8.76} & \underline{0.92} &- & - & - & - \\
  \multicolumn{2}{c}{StableVTON~\cite{kim2023stableviton}} 
  && 0.076 & 0.891 & 9.35 & 1.51 &- & - & - & - \\
  \multicolumn{2}{c}{GP-VTON*~\cite{xie2023gpvton}} 
  && 0.083 & 0.892 & 9.17 & 0.93 & 0.051 & 0.921 & 5.88 & 1.28  \\
 \multicolumn{2}{c}{LADI-VTON~\cite{morelli2023ladi}} 
  && 0.091 & 0.875 & 9.42 & 1.63 &0.072 & 0.902 & 6.94 & 2.33 \\
  \multicolumn{2}{c}{IDM~\cite{choi2024idm}} 
  && 0.078 & 0.881 & 9.12 & 1.03 & \underline{0.046} & \textbf{0.923} & \underline{5.32} & \underline{1.24} \\
    \cmidrule{1-2} \cmidrule{4-11}
  \multicolumn{2}{c}{\textbf{AnyFit(ours)}} 
  &&\underline{0.075} & \textbf{0.893} & \textbf{8.60} & \textbf{0.55}&\textbf{0.044} & \underline{0.904} & \textbf{4.51}& \textbf{0.48} \\  
  \bottomrule
\end{tabular}
\label{tab:viton}
\end{table*}

\begin{table*}[t]
\centering
\setlength{\tabcolsep}{2.6mm}
\begin{minipage}{0.52\textwidth}
  \centering
  \caption{Quantitative comparisons on the DressCode-multiple. The "Time" represents the inference time increase compared to its single-garment try-on.} 
  \fontsize{8pt}{9pt}\selectfont
  \begin{tabular}{l c c c c}
    \toprule
    Method & FID $\downarrow$ & KID $\downarrow$ & Time $\downarrow$ \\
    \cmidrule(lr){1-1}\cmidrule(lr){2-4}
    Paint-by-Example~\cite{yang2023paintbyexample} & 35.17 & 13.12 & 95\% \\
    IDM-2Stage~\cite{choi2024idm} & 21.47 & 7.85 & 93\% \\
  VTON-concat~\cite{chen2024wearanyway} & \underline{21.11} & \underline{7.30} & \textbf{8\%} \\
  
  \cmidrule(lr){1-1}\cmidrule(lr){2-4}
  \textbf{AnyFit (ours)} & \textbf{20.43} & \textbf{7.10} & \underline{9\%} \\
    \bottomrule
  \end{tabular}
  \label{tab:multi}
\end{minipage}%
\hfill
\begin{minipage}{0.46\textwidth}
  \centering
  \caption{Comparisons on proprietary dataset. \emph{AnyFit (xxx)} is trained only on \emph{xxx} dataset.} 
  \fontsize{8pt}{9pt}\selectfont
  \begin{tabular}{l c c c c }
    \toprule
    \multicolumn{2}{c}{Method}                              
    & & FID $\downarrow$ & KID $\downarrow$  \\
      \cmidrule{1-2} \cmidrule{4-5} 
      \multicolumn{2}{c}{LADI-VTON~\cite{morelli2023ladi}} 
    && 52.24 & 6.51 \\
    \multicolumn{2}{c}{DCI-VTON~\cite{gou2023dci}} 
    && 57.96 & 12.35  \\
    \multicolumn{2}{c}{StableVTON~\cite{kim2023stableviton}} 
    && 53.80 & 8.13 \\
    \multicolumn{2}{c}{IDM~\cite{choi2024idm}} 
    && 48.76 & 4.35 \\
      \cmidrule{1-2} \cmidrule{4-5}
    \multicolumn{2}{c}{\textbf{AnyFit (VITON-HD)}} 
    && \underline{46.95} & \underline{2.73} \\
    \multicolumn{2}{c}{\textbf{AnyFit (proprietary)}} 
    &&\textbf{43.97} & \textbf{0.69} \\  
    \bottomrule
  \end{tabular}
  \label{tab:wild}
\end{minipage}
\end{table*}

\subsection{Qualitative results}

\noindent \textbf{Single-garment try-on.}
Fig.\ref{fig:main_viton} and \ref{fig:main_wild} provide a qualitative comparison between AnyFit and the baselines on VITON-HD, the more challenging proprietary and in-the-wild data, covering open-garment and layering rendering scenarios. For a fair comparison with the baselines, we include results of AnyFit trained on VITON-HD. AnyFit excels in retaining intricate pattern details, owing to the effective collaboration between HydraNet and the IP-Adapter. It also maintains the correct silhouette of the clothing at a semantic level. This suggests that, through Mask Boost, AnyFit enhances the recollection of the original shape of the clothing, while other models, influenced by the mask, tend to generate incorrect appearances. The Prior Model Evolution further strengthens the texture representation of the apparel. Notably, when trained on the proprietary dataset, AnyFit automatically fills in inner garments or unzips clothing based on posture, a capability absent in the version trained on VITON-HD due to the lack of such training data.

\noindent \textbf{Multi-garment try-on.}
Fig.~\ref{fig:multi} offers a qualitative comparison for multi-garment try-ons using the compiled DressCode-multiple dataset. Firstly, AnyFit demonstrates high-fidelity cloth preservation. Importantly, thanks to the distinct and individual Hydra-Blocks situated in different condition branches, AnyFit accurately depicts the demarcation between the upper and lower garments, showcasing a reasonable transition at the interconnection. In contrast, VTON-concat mishandles the relative clothing sizes after concatenation, leading to garment distortion and blurring. Meanwhile, IDM-2Stage faces artifacts at the juncture of the upper and lower garments, because it obscures parts of one garment while trying on another. Remarkably, despite training with one garment presented as a flat lay image and the other as a warped cloth cropped from a person image, AnyFit remains  strikingly robust when faced with both garments presented as flat lays during inference.

\subsection{Quantitative results}
As indicated in Tab.~\ref{tab:viton} ~\ref{tab:multi} ~\ref{tab:wild}, extensive experiments conducted on VITON-HD~\cite{choi2021vitonhd}, DressCode~\cite{morelli2022DressCode}, the proprietary dataset, and DressCode-multiple consistently prove that AnyFit significantly surpasses all baselines. This confirms AnyFit's capability to deliver superior try-on quality in both single-garment and multi-garment tasks across various scenes. Moreover, we note that AnyFit shows considerable improvement in unpaired settings in terms of the FID and KID metrics, demonstrating our model's robustness for cross-category try-ons. For more results, please refer to the Appendix~\ref{appendix:exp_more}.

\subsection{Ablation study}

\begin{figure}[t]
  \centering
\includegraphics[width=1.0\textwidth]{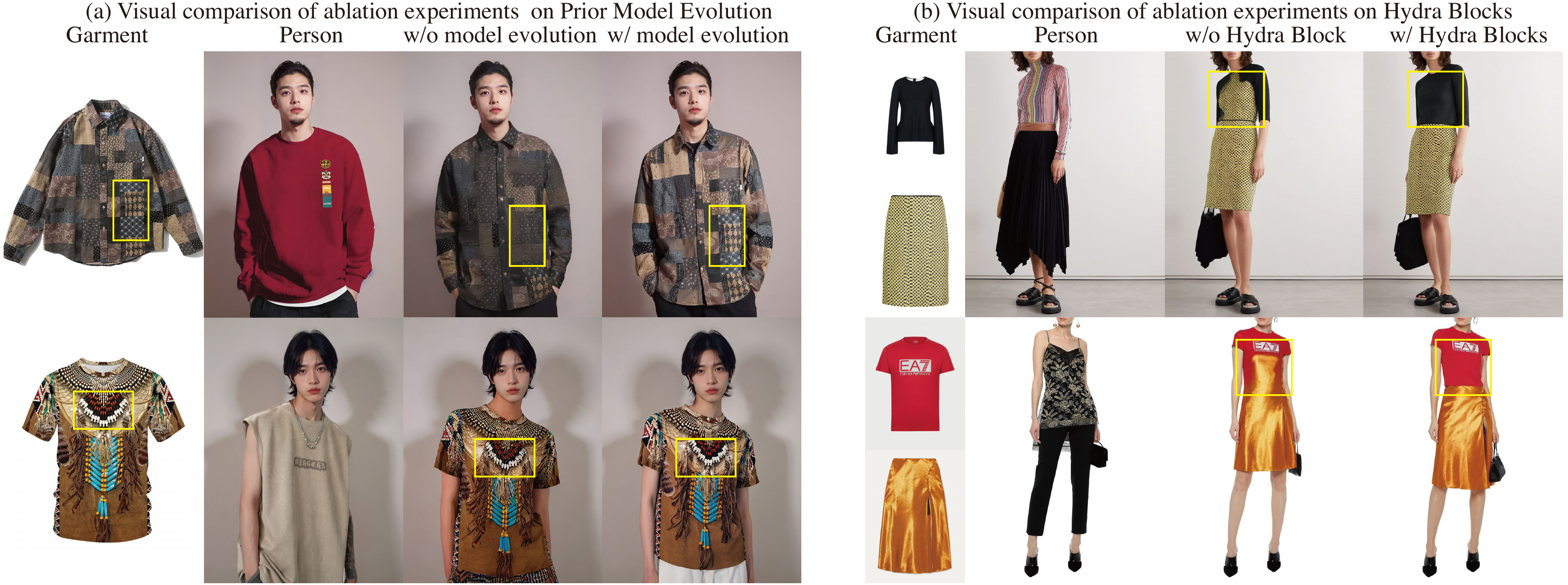}
  \caption{Visual ablation study. Without Prior Model Evolution, AnyFit suffers reduced fabric detail and less realistic textures. While Hydra Blocks improve intersections of upper and lower garments.} 
  \label{fig:ablation}
\end{figure}

\noindent \textbf{Hydra Blocks.} 
To validate our proposed Hydra Blocks, we directly employ a singular conditioned HydraNet (which degenerates to ReferenceNet~\cite{hu2023animate} actually) as the baseline "w/o Hydra Block" to encode both the top and bottom garment conditions concurrently, and then concatenate them into MainNet. As illustrated in Tab.~\ref{tab:ablation}, Fig.~\ref{fig:ablation} and ~\ref{fig:multi_ablation_supp}, a model lacking the Hydra Block tends to produce artifacts at the junction of the top and bottom garments. Such models also frequently allow the features of one garment to influence the other, leading to incorrect clothing styles. However, with the introduction of the Hydra Block, AnyFit consistently exhibits more stable results.

\noindent \textbf{Prior Model Evolution.}
We qualitatively demonstrate the effects of the Prior Model Evolution in Fig.~\ref{fig:cfg_merge} and ~\ref{fig:adaptive_mask} (a). The SDXL-evolved model reduces artifacts and enhances robustness significantly, while outputs without Prior Model Evolution typically feature oversaturated colors as well as lighting and shadows that do not harmonize with the background. The gradual enhancement of model capabilities is visualized in Fig.~\ref{fig:adaptive_mask}(c). We also empirically and quantitatively validate the effectiveness of the Prior Model Evolution strategy after training on the Virtual Try-On (VTON) task in Fig.~\ref{fig:ablation} and Tab.~\ref{tab:ablation}. The Prior Model Evolution, by improving the model's initial capabilities, lessens the difficulty of learning and fosters a dramatic improvement in outfitting capacity and logo fidelity.

\noindent \textbf{Adaptive Mask Boost.}
We illustratively showcase the issues of information leakage and mask reliance found in previous methods in Fig.~\ref{fig:adaptive_mask} (b) and Fig.~\ref{fig:parsing_leaky}. Additionally, we empirically and quantitatively validate the effectiveness of the Adaptive Mask Boost strategy in Table~\ref{tab:ablation} and Fig.~\ref{fig:parsing_leaky}. This strategy significantly heightens the model's robustness towards different categories of clothing, enabling the autonomous determination of appropriate garment length rather than relying on masks. Furthermore, we manually adjust the aspect ratios $\sigma$ in Fig.~\ref{fig:adaptive_mask} (d), which demonstrates the positive impact of adaptive elongation during inference. More ablation studies are detailed in the Appendix~\ref{appendix:ablation_more}.

\begin{table*}[t]
\centering
\caption{Quantitative ablation study.} 
\fontsize{8pt}{9pt}\selectfont
\begin{tabular}{llcccc}
\toprule
Dataset                & Method                  & LPIPS $\downarrow$ & SSIM $\uparrow$ & FID $\downarrow$ & KID $\downarrow$ \\ 
\cmidrule(lr){1-2}\cmidrule(lr){3-6}
\multirow{3}{*}{DressCode-multiple} 
                       & - w/o Hydra Blocks          & -          & -        & 22.48      & 8.02   \\
                       & - w/o Prior Model Evolution & -       & -           & 21.35      & 7.58  \\    
                       & \textbf{Full AnyFit}    & -    & -     & \textbf{20.43}      & \textbf{7.10}    \\
  \cmidrule(lr){1-2}\cmidrule(lr){3-6}
\multirow{3}{*}{the proprietary dataset}  
                       & - w/o Adaptive Mask Boost   & 0.183 & \textbf{0.748}  & 44.75  & 1.44      \\
                       & - w/o Prior Model Evolution & 0.192 & 0.740  & 45.01  & 1.52  \\ 
                       & \textbf{Full AnyFit}  & \textbf{0.181}  & 0.743  & \textbf{43.97} & \textbf{0.69}  \\
\bottomrule
\end{tabular}
\label{tab:ablation}
\end{table*}

\section{Conclusion}
We introduce AnyFit, a novel and robust VTON pipeline suitable for any combination of attire across any imaginable scenario, offering a revolutionary leap in realistic try-on effects. To support multi-garment try-ons, AnyFit constructs HydraNet with lightweight and scalable parallelized attention that facilitates the feature injection of multiple garments. Observing artifacts in real-world scenarios, we evolve its potential by synthesizing the residuals of multiple models, as well as implementing a mask region boost strategy. Comprehensive experiments on high-resolution benchmarks and real-world data have demonstrated that AnyFit significantly surpasses all baselines by a large gap.

\noindent \textbf{Broader impacts.}
With the ability to synthesize images, arises the risk that AnyFit might be used for inappropriate purposes such as producing media that breaches intellectual property rights or privacy norms. Because of these risks, we strongly advocate for the conscientious use of this technology.

\noindent\textbf{Limitation and future work.}
Our approach exhibits excellent performance in single-garment and multi-garment virtual try-on applications. However, it still faces some limitations. Firstly, it shares the shortcomings of large text-image models, sometimes showing instability in generating hands with complex structures. Secondly, our model offers initial but not yet fully mature text control capabilities (for details, please refer to the Appendix~\ref{appendix:ablation_more}), providing opportunities for future enhancements.

\clearpage  

\bibliographystyle{splncs04}
\bibliography{main}

\begin{thebibliography}{10}
\providecommand{\url}[1]{\texttt{#1}}
\providecommand{\urlprefix}{URL }
\providecommand{\doi}[1]{https://doi.org/#1}

\bibitem{worse_inpainting}
Stable diffusion xl shows worse inpainting. \url{https://www.reddit.com/r/StableDiffusion/comments/166bz7b/sdxl_base_model_for_inpainting_way_worse_than_15/ }

\bibitem{ModelMerge}
Discussions about model synthesis. \url{https://www.reddit.com/r/StableDiffusion/comments/zcby0o/you_can_now_merge_inpainting_and_regular_models/}

\bibitem{dreamshaperxl}
Dreamshaper xl. \url{https://civitai.com/models/112902/dreamshaper-xl}

\bibitem{sdxl_inp}
Sdxl-inpainting-0.1. \url{https://huggingface.co/diffusers/stable-diffusion-xl-1.0-inpainting-0.1}

\bibitem{deepspeed}
Deepspeed. \url{https://github.com/microsoft/DeepSpeed}

\bibitem{bai2022sdafn}
Bai, S., Zhou, H., Li, Z., Zhou, C., Yang, H.: Single stage virtual try-on via deformable attention flows. In: European Conference on Computer Vision (2022)

\bibitem{bookstein1989tps}
Bookstein, F.L.: Principal warps: Thin-plate splines and the decomposition of deformations. IEEE Transactions on pattern analysis and machine intelligence  (1989)

\bibitem{cao2017openpose}
Cao, Z., Simon, T., Wei, S.E., Sheikh, Y.: Realtime multi-person 2d pose estimation using part affinity fields. In: Proceedings of the IEEE conference on computer vision and pattern recognition (2017)

\bibitem{chang2023magicdance}
Chang, D., Shi, Y., Gao, Q., Fu, J., Xu, H., Song, G., Yan, Q., Yang, X., Soleymani, M.: Magicdance: Realistic human dance video generation with motions \& facial expressions transfer. arXiv preprint arXiv:2311.12052  (2023)

\bibitem{chen2024wearanyway}
Chen, M., Chen, X., Zhai, Z., Ju, C., Hong, X., Lan, J., Xiao, S.: Wear-any-way: Manipulable virtual try-on via sparse correspondence alignment. arXiv preprint  (2024)

\bibitem{chen2023anydoor}
Chen, X., Huang, L., Liu, Y., Shen, Y., Zhao, D., Zhao, H.: Anydoor: Zero-shot object-level image customization. arXiv preprint arXiv:2307.09481  (2023)

\bibitem{choi2021vitonhd}
Choi, S., Park, S., Lee, M., Choo, J.: Viton-hd: High-resolution virtual try-on via misalignment-aware normalization. In: Proceedings of the IEEE/CVF conference on computer vision and pattern recognition (2021)

\bibitem{choi2024idm}
Choi, Y., Kwak, S., Lee, K., Choi, H., Shin, J.: Improving diffusion models for virtual try-on. arXiv preprint arXiv:2403.05139  (2024)

\bibitem{chopra2021zflow}
Chopra, A., Jain, R., Hemani, M., Krishnamurthy, B.: Zflow: Gated appearance flow-based virtual try-on with 3d priors. In: Proceedings of the IEEE/CVF International Conference on Computer Vision (2021)

\bibitem{cui2021dressing}
Cui, A., McKee, D., Lazebnik, S.: Dressing in order: Recurrent person image generation for pose transfer, virtual try-on and outfit editing. In: Proceedings of the IEEE/CVF international conference on computer vision (2021)

\bibitem{morelli2022DressCode}
Davide, M., Matteo, F., Marcella, C., Federico, L., Fabio, C., Rita, C.: Dress code: High-resolution multi-category virtual try-on. In: Proceedings of the IEEE/CVF Conference on Computer Vision and Pattern Recognition (2022)

\bibitem{dong2020fashion}
Dong, H., Liang, X., Zhang, Y., Zhang, X., Shen, X., Xie, Z., Wu, B., Yin, J.: Fashion editing with adversarial parsing learning. In: Proceedings of the IEEE/CVF conference on computer vision and pattern recognition. pp. 8120--8128 (2020)

\bibitem{Ge2021PFAFN}
Ge, Y., Song, Y., Zhang, R., Ge, C., Liu, W., Luo, P.: Parser-free virtual try-on via distilling appearance flows. In: Proceedings of the IEEE/CVF conference on computer vision and pattern recognition (2021)

\bibitem{goodfellow2014GAN}
Goodfellow, I., Pouget-Abadie, J., Mirza, M., Xu, B., Warde-Farley, D., Ozair, S., Courville, A., Bengio, Y.: Generative adversarial nets. Advances in neural information processing systems  (2014)

\bibitem{gou2023dci}
Gou, J., Sun, S., Zhang, J., Si, J., Qian, C., Zhang, L.: Taming the power of diffusion models for high-quality virtual try-on with appearance flow. In: Proceedings of the 31st ACM International Conference on Multimedia (2023)

\bibitem{guler2018densepose}
G{\"u}ler, R.A., Neverova, N., Kokkinos, I.: Densepose: Dense human pose estimation in the wild. In: Proceedings of the IEEE conference on computer vision and pattern recognition (2018)

\bibitem{han2019clothflow}
Han, X., Hu, X., Huang, W., Scott, M.R.: Clothflow: A flow-based model for clothed person generation. In: Proceedings of the IEEE/CVF international conference on computer vision (2019)

\bibitem{han2018viton}
Han, X., Wu, Z., Wu, Z., Yu, R., Davis, L.S.: Viton: An image-based virtual try-on network. In: Proceedings of the IEEE conference on computer vision and pattern recognition (2018)

\bibitem{he2022fs_vton}
He, S., Song, Y.Z., Xiang, T.: Style-based global appearance flow for virtual try-on. In: Proceedings of the IEEE/CVF Conference on Computer Vision and Pattern Recognition (2022)

\bibitem{ho2020ddpm}
Ho, J., Jain, A., Abbeel, P.: Denoising diffusion probabilistic models. Advances in neural information processing systems  (2020)

\bibitem{hu2023animate}
Hu, L., Gao, X., Zhang, P., Sun, K., Zhang, B., Bo, L.: Animate anyone: Consistent and controllable image-to-video synthesis for character animation. arXiv preprint arXiv:2311.17117  (2023)

\bibitem{huang2023composer}
Huang, L., Chen, D., Liu, Y., Shen, Y., Zhao, D., Zhou, J.: Composer: Creative and controllable image synthesis with composable conditions. arXiv preprint arXiv:2302.09778  (2023)

\bibitem{jin2023sssegmenation}
Jin, Z.: Sssegmenation: An open source supervised semantic segmentation toolbox based on pytorch. arXiv preprint arXiv:2305.17091  (2023)

\bibitem{jin2024idrnet}
Jin, Z., Hu, X., Zhu, L., Song, L., Yuan, L., Yu, L.: Idrnet: Intervention-driven relation network for semantic segmentation. Advances in Neural Information Processing Systems  \textbf{36} (2024)

\bibitem{jo2019scfegan}
Jo, Y., Park, J.: Sc-fegan: Face editing generative adversarial network with user's sketch and color. In: Proceedings of the IEEE/CVF international conference on computer vision. pp. 1745--1753 (2019)

\bibitem{kim2023stableviton}
Kim, J., Gu, G., Park, M., Park, S., Choo, J.: Stableviton: Learning semantic correspondence with latent diffusion model for virtual try-on. In: Proceedings of the IEEE/CVF Conference on Computer Vision and Pattern Recognition (2024)

\bibitem{kingma2013vae}
Kingma, D.P., Welling, M.: Auto-encoding variational bayes. arXiv preprint arXiv:1312.6114  (2013)

\bibitem{kirillov2023sam}
Kirillov, A., Mintun, E., Ravi, N., Mao, H., Rolland, C., Gustafson, L., Xiao, T., Whitehead, S., Berg, A.C., Lo, W.Y., et~al.: Segment anything. arXiv preprint arXiv:2304.02643  (2023)

\bibitem{lee2020maskgan}
Lee, C.H., Liu, Z., Wu, L., Luo, P.: Maskgan: Towards diverse and interactive facial image manipulation. In: Proceedings of the IEEE/CVF conference on computer vision and pattern recognition. pp. 5549--5558 (2020)

\bibitem{lee2022hrviton}
Lee, S., Gu, G., Park, S., Choi, S., Choo, J.: High-resolution virtual try-on with misalignment and occlusion-handled conditions. In: European Conference on Computer Vision (2022)

\bibitem{lin2017fpn}
Lin, T.Y., Doll{\'a}r, P., Girshick, R., He, K., Hariharan, B., Belongie, S.: Feature pyramid networks for object detection. In: Proceedings of the IEEE conference on computer vision and pattern recognition (2017)

\bibitem{loshchilov2018adamw}
Loshchilov, I., Hutter, F.: Fixing weight decay regularization in adam. arXiv  (2018)

\bibitem{morelli2023ladi}
Morelli, D., Baldrati, A., Cartella, G., Cornia, M., Bertini, M., Cucchiara, R.: {LaDI-VTON: Latent Diffusion Textual-Inversion Enhanced Virtual Try-On}. In: Proceedings of the ACM International Conference on Multimedia (2023)

\bibitem{nichol2021glide}
Nichol, A., Dhariwal, P., Ramesh, A., Shyam, P., Mishkin, P., McGrew, B., Sutskever, I., Chen, M.: Glide: Towards photorealistic image generation and editing with text-guided diffusion models. arXiv preprint arXiv:2112.10741  (2021)

\bibitem{parmar2021cleanfid}
Parmar, G., Zhang, R., Zhu, J.Y.: On aliased resizing and surprising subtleties in gan evaluation. In: Proceedings of the IEEE/CVF Conference on Computer Vision and Pattern Recognition (2022)

\bibitem{podell2023sdxl}
Podell, D., English, Z., Lacey, K., Blattmann, A., Dockhorn, T., M{\"u}ller, J., Penna, J., Rombach, R.: Sdxl: Improving latent diffusion models for high-resolution image synthesis. arXiv preprint arXiv:2307.01952  (2023)

\bibitem{radford2021clip}
Radford, A., Kim, J.W., Hallacy, C., Ramesh, A., Goh, G., Agarwal, S., Sastry, G., Askell, A., Mishkin, P., Clark, J., et~al.: Learning transferable visual models from natural language supervision. In: International conference on machine learning (2021)

\bibitem{rombach2022ldm}
Rombach, R., Blattmann, A., Lorenz, D., Esser, P., Ommer, B.: High-resolution image synthesis with latent diffusion models. In: Proceedings of the IEEE/CVF conference on computer vision and pattern recognition (2022)

\bibitem{song2020ddim}
Song, J., Meng, C., Ermon, S.: Denoising diffusion implicit models. arXiv preprint arXiv:2010.02502  (2020)

\bibitem{sutherland2018kid}
Sutherland, J., Arbel, M., Gretton, A.: Demystifying mmd gans. In: International Conference for Learning Representations (2018)

\bibitem{vaswani2017attention}
Vaswani, A., Shazeer, N., Parmar, N., Uszkoreit, J., Jones, L., Gomez, A.N., Kaiser, {\L}., Polosukhin, I.: Attention is all you need. Advances in neural information processing systems  (2017)

\bibitem{wang2018toward}
Wang, B., Zheng, H., Liang, X., Chen, Y., Lin, L., Yang, M.: Toward characteristic-preserving image-based virtual try-on network. In: Proceedings of the European conference on computer vision (ECCV) (2018)

\bibitem{wang2004ssim}
Wang, Z., Bovik, A.C., Sheikh, H.R., Simoncelli, E.P.: Image quality assessment: from error visibility to structural similarity. IEEE transactions on image processing  (2004)

\bibitem{xie2023gpvton}
Xie, Z., Huang, Z., Dong, X., Zhao, F., Dong, H., Zhang, X., Zhu, F., Liang, X.: Gp-vton: Towards general purpose virtual try-on via collaborative local-flow global-parsing learning. In: Proceedings of the IEEE/CVF Conference on Computer Vision and Pattern Recognition (2023)

\bibitem{xu2024ootdiffusion}
Xu, Y., Gu, T., Chen, W., Chen, C.: Ootdiffusion: Outfitting fusion based latent diffusion for controllable virtual try-on. arXiv preprint arXiv:2403.01779  (2024)

\bibitem{yang2023paintbyexample}
Yang, B., Gu, S., Zhang, B., Zhang, T., Chen, X., Sun, X., Chen, D., Wen, F.: Paint by example: Exemplar-based image editing with diffusion models. In: Proceedings of the IEEE/CVF Conference on Computer Vision and Pattern Recognition (2023)

\bibitem{yang2020acgpn}
Yang, H., Zhang, R., Guo, X., Liu, W., Zuo, W., Luo, P.: Towards photo-realistic virtual try-on by adaptively generating-preserving image content. In: Proceedings of the IEEE/CVF conference on computer vision and pattern recognition (2020)

\bibitem{yang2023effective}
Yang, Z., Zeng, A., Yuan, C., Li, Y.: Effective whole-body pose estimation with two-stages distillation. In: Proceedings of the IEEE/CVF International Conference on Computer Vision. pp. 4210--4220 (2023)

\bibitem{ye2023ipadapter}
Ye, H., Zhang, J., Liu, S., Han, X., Yang, W.: Ip-adapter: Text compatible image prompt adapter for text-to-image diffusion models. arXiv preprint arXiv:2308.06721  (2023)

\bibitem{zhang2023controlnet}
Zhang, L., Rao, A., Agrawala, M.: Adding conditional control to text-to-image diffusion models. In: Proceedings of the IEEE/CVF International Conference on Computer Vision. pp. 3836--3847 (2023)

\bibitem{zhang2018perceptual}
Zhang, R., Isola, P., Efros, A.A., Shechtman, E., Wang, O.: The unreasonable effectiveness of deep features as a perceptual metric. In: Proceedings of the IEEE conference on computer vision and pattern recognition (2018)

\bibitem{zhang2024mmtryon}
Zhang, X., Lin, E., Li, X., Luo, Y., Kampffmeyer, M., Dong, X., Liang, X.: Mmtryon: Multi-modal multi-reference control for high-quality fashion generation. arXiv preprint arXiv:2405.00448  (2024)

\bibitem{zhu2023tryondiffusion}
Zhu, L., Yang, D., Zhu, T., Reda, F., Chan, W., Saharia, C., Norouzi, M., Kemelmacher-Shlizerman, I.: Tryondiffusion: A tale of two unets. In: Proceedings of the IEEE/CVF Conference on Computer Vision and Pattern Recognition (2023)

\end{thebibliography}

\newpage
\appendix

In Section~\ref{appendix:ablation_more}, we provide more ablation studies about the role of text prompt in AnyFit, Hydra Block, Adaptive Mask Boost strategy and Prior Model Evolution. 
In Section~\ref{appendix:datasets}, we provide additional details about the proprietary dataset and the multi-garment VTON dataset.
In Section~\ref{appendix:supplement_method}, we provide additional details about discrete greedy algorithm during Prior Model Evolution, and preliminary knowledge about stable diffusion.
In Section~\ref{appendix:exp_details}, we provide additional details about experimental  setup, including data augmentation and outfitting dropout.
Lastly, in Section~\ref{appendix:exp_more}, we present a multitude of AnyFit generated images, including single-garment visual try-ons and multi-garment virtual try-ons, with additional results displayed in challenging scenarios.

\section{More ablation study}
\label{appendix:ablation_more}

\subsection{The role of text prompt in AnyFit}
\begin{figure}[ht]
  \centering
\includegraphics[width=1.0\textwidth]{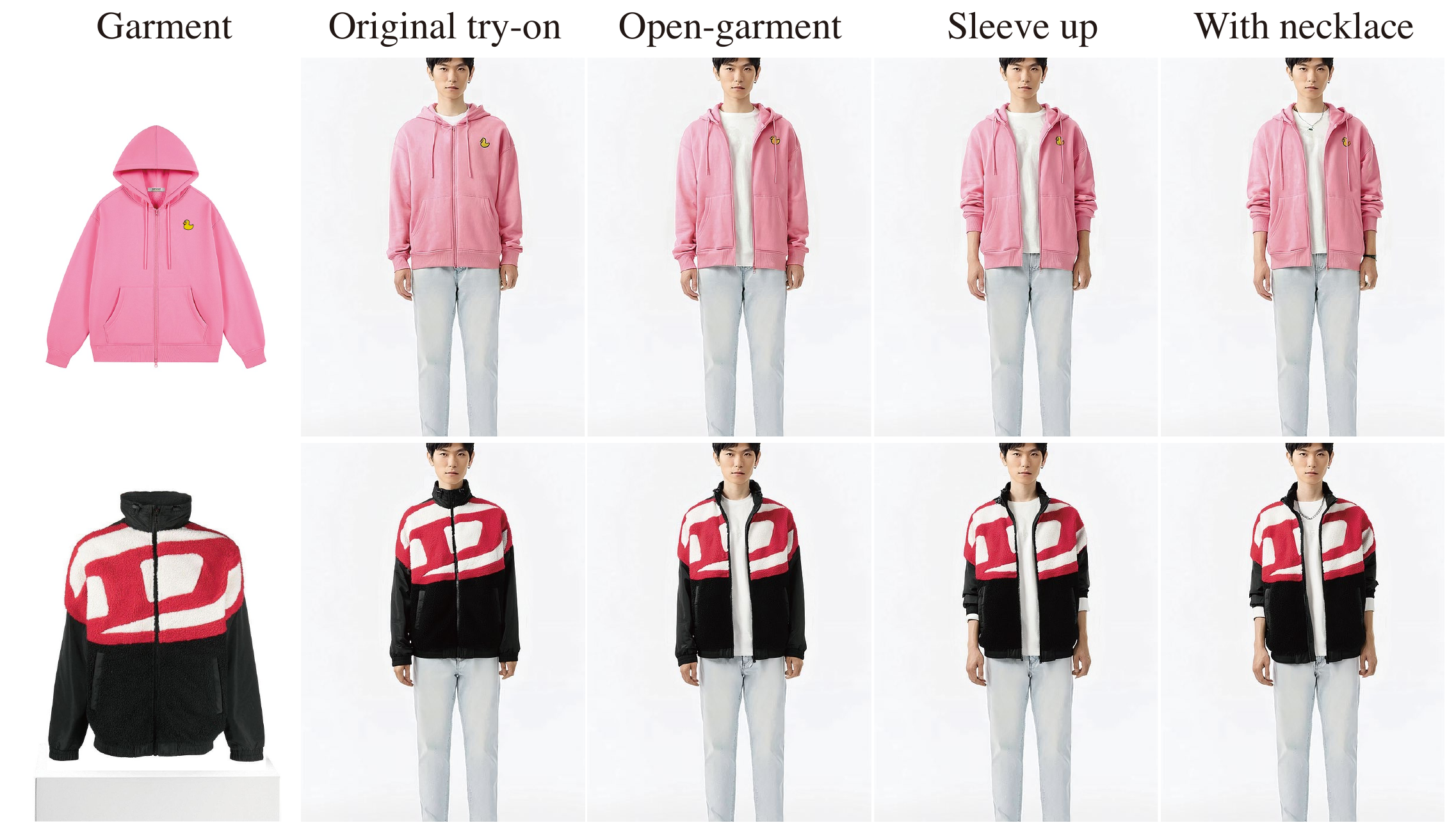}
  \caption{By adjusting the prompt, AnyFit is able to achieve variations in VTON apparel styles.} 
  \label{fig:text_control_supp}
\end{figure}

In fact, we have discovered that text plays a certain role in controlling the overall try-on style. As show in Fig.~\ref{fig:text_control_supp}, by adjusting the prompt, AnyFit is able to achieve variations in Virtual Try-On (VTON) apparel styles. However, this form of control is unstable, leaving space for further exploration.

\begin{figure}[ht]
  \centering
\includegraphics[width=1.0\textwidth]{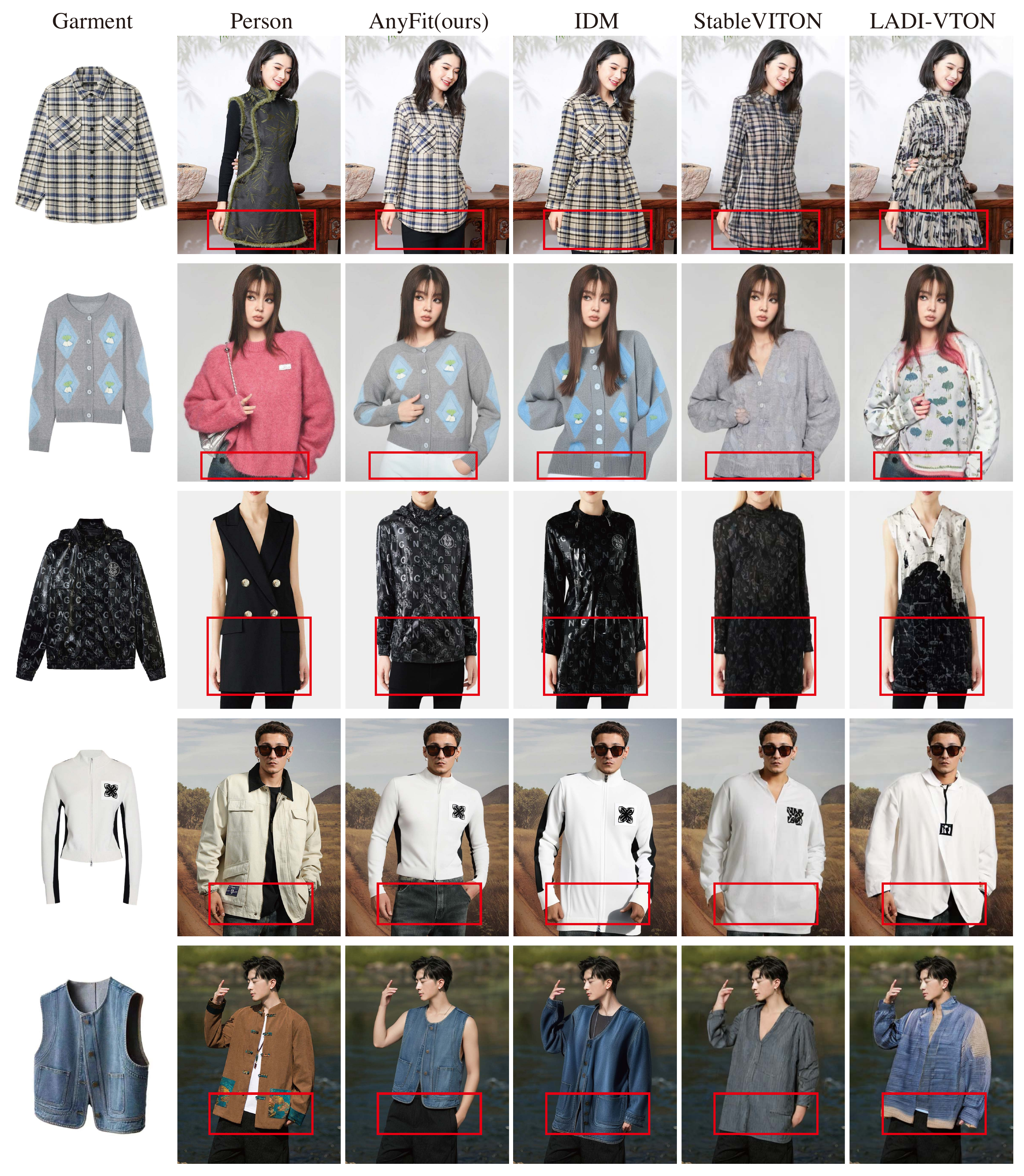}
  \caption{Visual validation of the role of parsing-free mask augmentation during training within Adaptive Mask Boost strategy. AnyFit is only trained on VITON-HD for a fair comparison. Previous methods often suffer from leakage of clothing information, where the model tends to generate garments that fill the entire mask area, while AnyFit autonomously determines the correct length of the garments, producing attractive try-on results.} 
  \label{fig:parsing_leaky}
\end{figure}

\begin{figure}[ht]
  \centering
\includegraphics[width=1.0\textwidth]{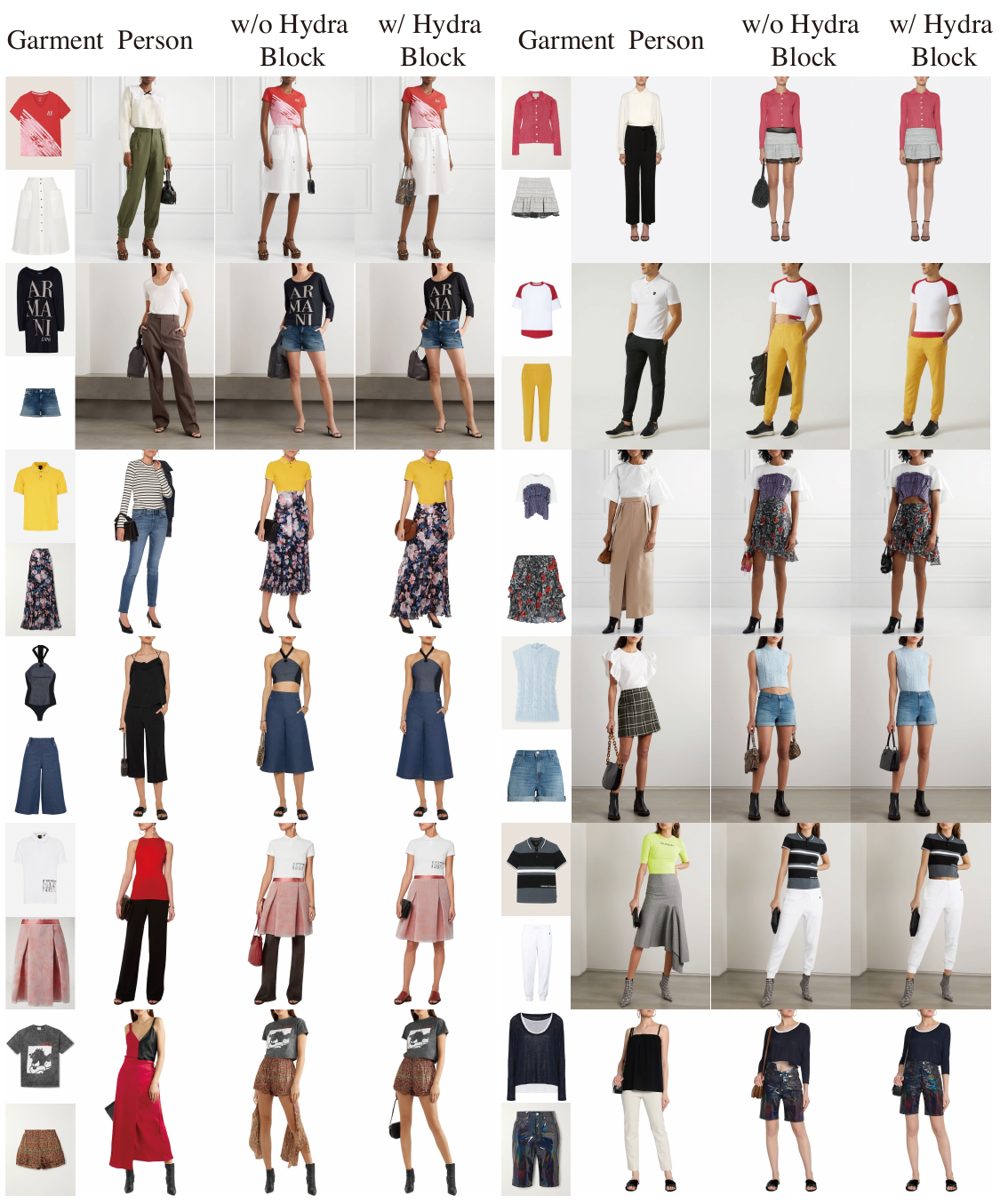}
  \caption{Visual comparisons on the DressCode-multiple. Model lacking the Hydra Block is more prone to producing artifacts at the junction of the top and bottom garments.} 
  \label{fig:multi_ablation_supp}
\end{figure}

\subsection{More ablation study about Adaptive Mask Boost strategy}

The Adaptive Mask Boost strategy primarily comprises mask augmentation during training and adaptive elongation during inference. The role of adaptive elongation has been thoroughly illustrated in Fig.~\ref{fig:adaptive_mask} in the main body, where it addresses the challenge of try-ons for long garments. Here, we predominantly discuss the function of parsing-free mask augmentation during training.

Specific to the implementation, during training, the agnostic mask is extracted solely using OpenPose body joint detections to occlude the original clothing, without leveraging human parsing, and random elongation of the mask is performed. Consequently, the shape of the mask is correlated only with the human pose and not with the clothing. \textbf{The significance of this procedure lies in eliminating the possibility of the mask leaking clothing information, ensuring that the model cannot cheat from the shape of the mask}. From Fig.~\ref{fig:parsing_leaky}, we observe that previous models often suffer from leakage of clothing information, where the models tend to generate garments that fill the entire mask areas. In real-world inference scenarios, it is challenging to provide an entirely accurate mask region. In contrast, our model is free from this predicament. Through mask augmentation during training, AnyFit autonomously determines the correct length of the garments, producing attractive try-on results.

\subsection{More ablation study about Hydra Block}

In Fig.~\ref{fig:multi_ablation_supp}, we present additional visual results to validate the effectiveness of the Hydra Block. "w/ Hydra Block" represents our fully proposed AnyFit, whereas the version "w/o Hydra Block" omits the Hydra Encoding Block as well as the positional embeddings within the Hydra Fusion. It directly employs a singular conditioned HydraNet (which, in this scenario, degenerates to ReferenceNet) to encode both the top and bottom garment conditions concurrently, and then injects them into MainNet.

As illustrated in Fig.~\ref{fig:multi_ablation_supp}, model lacking the Hydra Block is more prone to producing artifacts at the junction of the top and bottom garments. They also tend to allow the features of the bottom garment to influence the top, resulting in incorrect clothing styles, such as overly long tops or erroneous stripe patterns. After equipped with the Hydra Block, AnyFit noticeably exhibits more stable results.

\subsection{More ablation study about Prior Model Evolution}

In the main body of the paper, we have demonstrated the effects of Prior Model Evolution. However, we also notice that the capabilities of large-scale text-to-image models are closely related to the value of their classifier-free guidance (CFG). For simplicity, in the main body of the paper, all images are generated using a CFG value of $3.0$ in Fig.~\ref{fig:adaptive_mask} when discussing the comparisons related to Prior Model Evolution. Here, we provide a more granular comparison concerning the presence or absence of the Prior Model Evolution strategy. The results presented here are all based on models directly initialized with synthetic weights, without any training.

\begin{figure}[t]
  \centering
\includegraphics[width=1.0\textwidth]{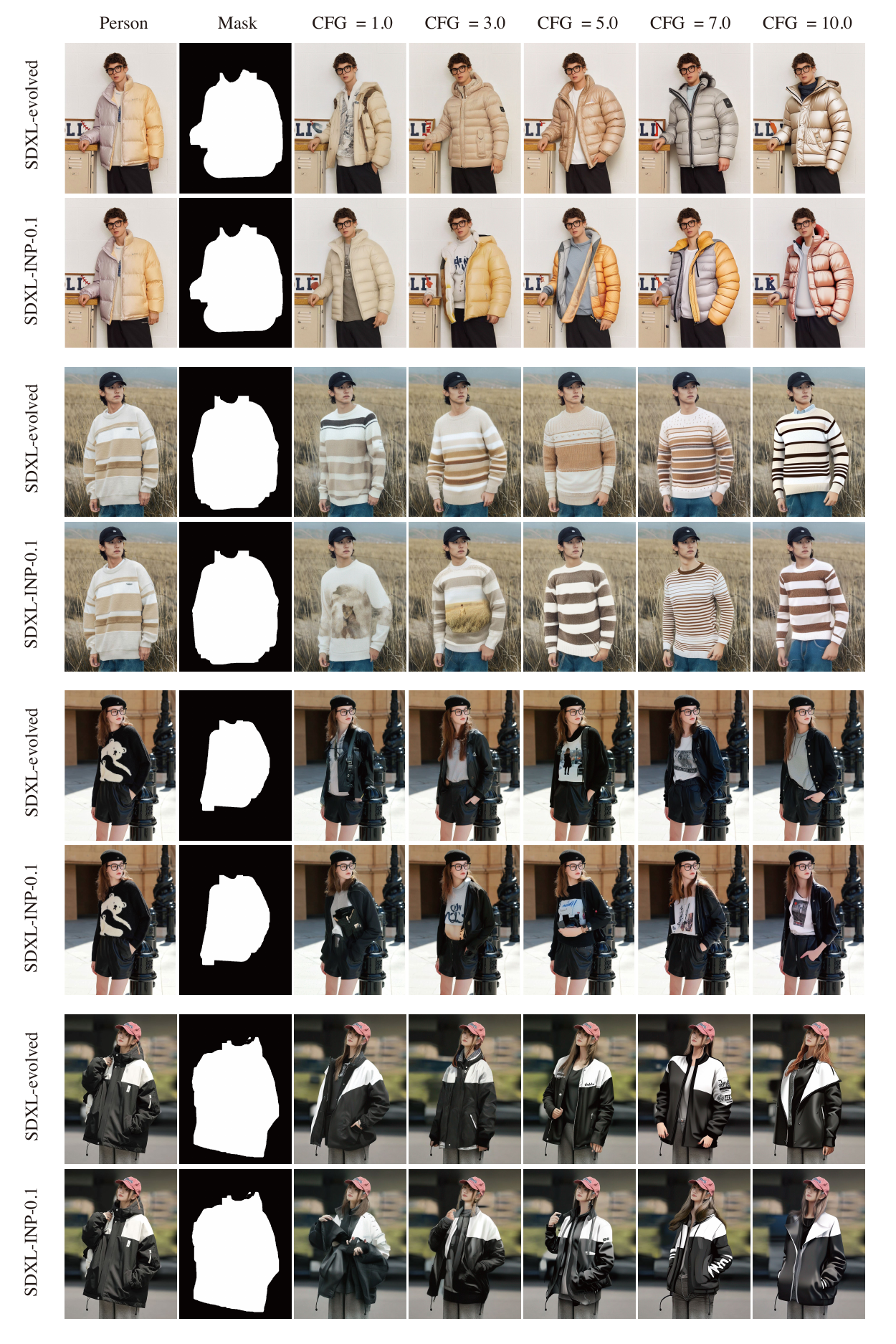}
  \caption{Visual validation of the role of Prior Model Evolution in various CFG weights without any training. Outputs without Prior Model Evolution typically feature oversaturated colors and the absence of detailed wrinkles, as well as lighting and shadows that do not harmonize with the background. Best viewed when zoomed in.} 
  \label{fig:cfg_merge}
\end{figure}

As depicted in Fig.~\ref{fig:cfg_merge}, an increase in the CFG value leads to a generalized oversaturation in the images. Within a reasonable range of CFG values, models incorporating Prior Model Evolution exhibit more realistic fabric textures and more plausible inpainting results. In contrast, outputs without Prior Model Evolution typically feature oversaturated colors and the absence of detailed wrinkles, as well as lighting and shadows that do not harmonize with the background. This illustrates the SDXL-inpainting-0.1~\cite{podell2023sdxl} model diminishes the exemplary text-to-image capabilities of the original SDXL-BASE model, resulting in more mediocre outcomes. We attribute this degradation to the disruption of the previously well-aligned correspondences between text and images during the inpainting pre-training phase. However, our model with Prior Model Evolution, significantly ameliorates this issue, enhancing its overall robustness. We actually find that \textbf{after a minimal amount of fine-tuning, the model's synthetic outputs become substantially more powerful}.

\section{Datasets}
\label{appendix:datasets}
In this section, we provide a detailed description of the proprietary dataset and the multi-garment VTON dataset, \emph{i.e.}, DressCode-multiple, which is constructed based on the publicly available DressCode~\cite{morelli2022DressCode} dataset.

\subsection{The proprietary dataset}

\begin{figure}[h]
  \centering
  \vspace{-5mm}
\includegraphics[width=1.0\textwidth]{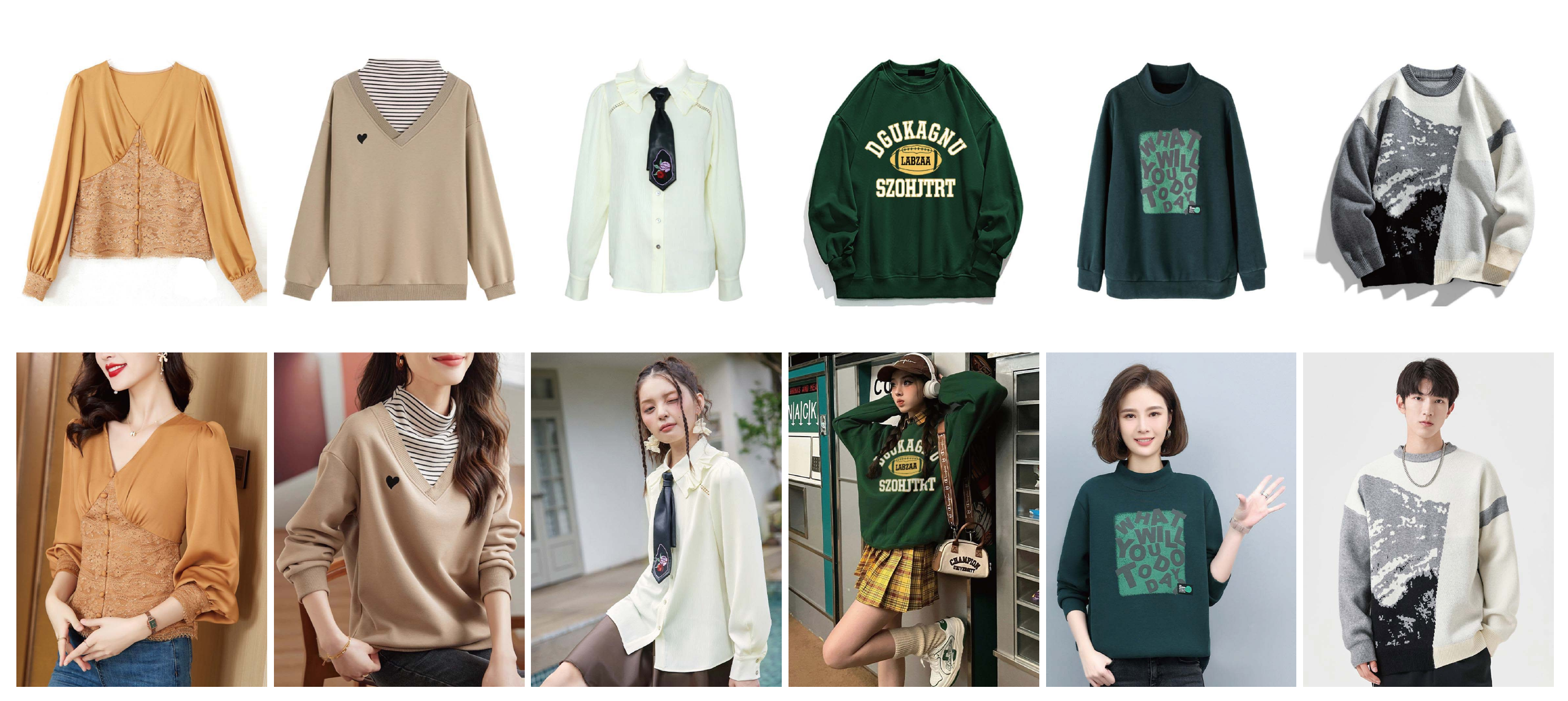}
\vspace{-5mm}
  \caption{Examples of the proprietary dataset. } 
  \vspace{-5mm}
  \label{fig:datashow_wild_supp}
\end{figure}

Our proprietary dataset comprises 50,602 training image pairs and 2,500 testing pairs. Each pair consists of a flat-laid garment image and a frontal upper body model image, with most model images featuring complex backgrounds. The proprietary dataset is collected from e-commerce websites, with attention paid to achieving a balanced distribution of clothing categories. We gathered apparel from 12 different categories, including both menswear and womenswear, across various seasons such as spring, summer, autumn, and winter. As illustrated in Fig.~\ref{fig:datashow_wild_supp}, the person images display diverse backgrounds. We standardized the image resolution to 1024x768 and conducted preprocessing on this basis to test different models. This preprocessing involved using Densepose~\cite{guler2018densepose} and OpenPose~\cite{cao2017openpose} to extract human features, employing the SAM~\cite{kirillov2023sam} model to extract the main garment body, and constructing the gnostic mask images, \emph{etc}.

\clearpage

\begin{figure}[h]
  \centering
\includegraphics[width=1.0\textwidth]{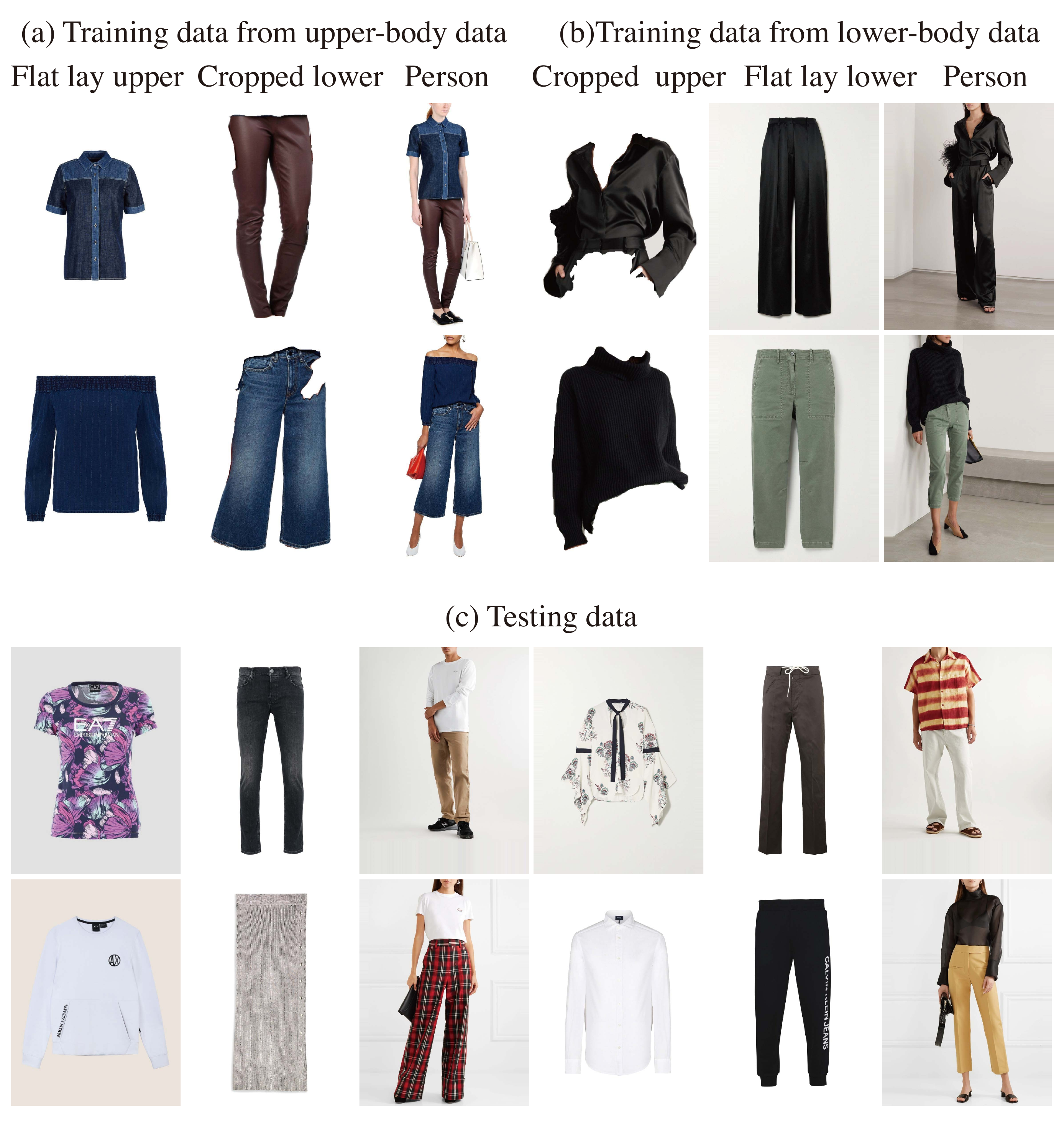}
  \caption{Examples of the DressCode-multiple dataset. } 
  \label{fig:datashow_multi_supp}
\end{figure}

\subsection{The DressCode-multiple dataset}

To facilitate research on multi-garment virtual try-on, we require a dataset composed of image triplets, each containing an upper garment image, a lower garment image, and a model image wearing the corresponding garments. However, obtaining such data with strict alignment is challenging. Leveraging the DressCode dataset~\cite{morelli2022DressCode}, which includes upper and lower garment data as well as full-body model images, we set out to construct the DressCode-multiple dataset, consisting of triplets, as illustrated in Fig.~\ref{fig:datashow_multi_supp}. Assuming we start with the upper garment data, where we already have a flat lay upper garment image and a model image wearing the corresponding upper garment, we use human parsing techniques~\cite{jin2023sssegmenation, jin2024idrnet} to roughly segment the lower garment portion and extract it from the model image to serve as the corresponding lower garment image. At this point, the triplet consists of (flat lay upper, cropped lower, model image). Similarly, triplets derived from the lower garment data result in a composition of (cropped upper, flat lay lower, model image).
Using this approach, we construct 24,314 publicly available upper-lower triplets by cropping garments from the 15,363 upper-body pairs and 8,951 lower-body pairs for training. For testing, we take the flat-laid garments from the upper and lower garment test sets, shuffle them, and combine them randomly, pairing them with the test models from the upper garment data to create (flat lay upper, flat lay lower, model image) configurations for unpaired, real-world multi-garment virtual try-on tests. The paired testing is not conducted due to the lack of ground truth triplets.

It is worth mentioning that our model has not encountered triplets in the form of (flat lay upper garment, flat lay lower garment, model image) during training. However, it has already learned the correct way to wear upper and lower garments through training on cropped images and demonstrates good robustness. We believe that if real triplets consisting of (flat lay upper garment, flat lay lower garment, model image) were available for training, the model would exhibit even better performance.

\section{Method supplement}
\label{appendix:supplement_method}

\subsection{Discrete greedy algorithm} 
\label{appendix:algorithm}

Our objective is to determine the optimal balancing coefficients $\alpha$ and $\beta$ to ensure that the initial weight $\mathbf{W}_{new}$ achieve the best evaluation performance, i.e. 
$$\mathop{\arg\min}_{(\alpha, \beta)\in [0,2]^2} f(\alpha, \beta) =  \Phi\left( \mathbf{W}_{base} + 
\alpha \cdot \left ( \mathbf{W}_{inp} - \mathbf{W}_{base} \right )
+ \beta \cdot  \left ( \mathbf{W}_{ds} - \mathbf{W}_{base} \right) \right)$$
Note that the function $\Phi$ is non-differentiable, which precludes the use of gradient-based optimization algorithms to find the minimum point.
Empirically, we find that the evaluation function $f$ exhibits monotonic or convex properties with respect to the balancing coefficients $(\alpha, \beta)$ in most regions. 
Therefore, we discretize the continuous domain $[0,2]^2$ into a grid with $\delta$ as the step size and design the following algorithm, inspired by the greedy method, to search for the optimal $(\alpha, \beta)$.

\begin{algorithm}
\caption{Discrete greedy algorithm}
\begin{algorithmic}[1]
\Require Evaluation function $f$, step size $\delta$.
\State Initialize $(\alpha, \beta) \gets (0.5, 0.5)$
\While{True}
    \State $f_{\text{current}} \gets f(\alpha, \beta)$
    \State $N \gets \{(\alpha + \delta, \beta), (\alpha - \delta, \beta), (\alpha, \beta + \delta), (\alpha, \beta - \delta)\}$ 
    \State $N \gets \{(\alpha', \beta') \in N \mid 0 \leq \alpha' \leq 2 \text{ and } 0 \leq \beta' \leq 2\}$ \Comment{Filter out-of-bound neighbors} 
    \State $F_N \gets \{f(\alpha', \beta') \mid (\alpha', \beta') \in N\}$ \Comment{Compute $f$ values for valid neighbors} 
    \If{$\min(F_N) \geq f_{\text{current}}$}
        \State \textbf{break}
    \Else
        \State $(\alpha, \beta) \gets \arg\min_{(\alpha', \beta') \in N} f(\alpha', \beta')$
    \EndIf
\EndWhile
\State \Return $(\alpha, \beta)$
\end{algorithmic}
\end{algorithm}

In this algorithm, we initialize $(\alpha, \beta)$ at $(0.5, 0.5)$. 
During each iteration, we calculate the evaluation value $f$ at the current point and four adjacent points in the directions up, down, left, and right. 
The algorithm then updates the current point to the one with the minimum evaluation value among these adjacent points. 
The termination condition of the algorithm is met when the evaluation values of all neighbors are greater than that of the current point. 
This method can be viewed as a discrete version of gradient descent method.
Please note that under the assumption that $f$ is a convex or monotonic function, the algorithm is guaranteed to converge to the global optimal solution within the discrete parameter space.

\subsection{Preliminary}
\label{appendix:preliminary}

\paragraph{Stable Diffusion.}
Our AnyFit is an extension of Stable Diffusion~\cite{rombach2022ldm}, which is one of the most commonly used latent diffusion models. Stable Diffusion employs a variational autoencoder~\cite{kingma2013vae} (VAE) that consists of an encoder $\mathcal{E}$ and a decoder $\mathcal{D}$ to enable image representations in the latent space. And a UNet $\epsilon_{\theta}$ is trained to denoise a Gaussian noise $\epsilon$ with a conditioning input encoded by a CLIP text encoder~\cite{radford2021clip} $\tau_{\theta}$. Given an image $\mathbf{x}$ and a text prompt $\mathbf{y}$, the training of the denoising UNet $\epsilon_{\theta}$ is performed by minimizing the following loss function:
\begin{equation}
    \mathcal{L}_{LDM} = \mathbb{E}_{\mathcal{E}(\mathbf{x}),\mathbf{y},\epsilon\sim\mathcal{N}(0, 1),t}\left[\lVert\epsilon - \epsilon_{\theta}(\mathbf{z}_t, t, \tau_{\theta}(\mathbf{y}))\rVert_2^2\right],
\end{equation}
where $t\in\{1,...,T\}$ denotes the time step of the forward diffusion process, and $\mathbf{z}_t$ is the encoded image $\mathcal{E}(\mathbf{x})$ with the added Gaussian noise $\epsilon\sim\mathcal{N}(0, 1)$ (\emph{i.e.}, the noise latent). Note that the conditioning input $\tau_{\theta}(\mathbf{y})$ is correlated with the denoising UNet by the cross-attention mechanism~\cite{vaswani2017attention}.

\section{Experimental details}
\label{appendix:exp_details}

\paragraph{Data augmentation.} 
We have implemented data augmentation techniques that could potentially enhance the model's generalization ability as well as its color accuracy performance. Specifically, the data augmentation operations include (a) horizontal flipping of images, (b) resizing garments and human figures through padding (up to 10\% of the image size), (c) randomly adjusting the image's hue within a range of -5 to +5, and (d) randomly adjusting the image's contrast within a specified range (between 0.8 and 1.2 times the original contrast). Each of these operations occurs independently with a 50\% probability. Moreover, these operations are simultaneously applied to both the garment and model images.

\paragraph{Outfitting dropout.} 
Following the OOTDiffusion approach, we applied Outfitting Dropout, which essentially acts as a form of image-conditioned classifier-free guidance. It enhances the contrast and sharpness of the generated images. Specifically, during the training process of our MainNet, we randomly drop the input garment latent as $E(g) = \varnothing$, where $\varnothing \in \mathbb{R}^{4\times h\times w}$ refers to an all-zero latent. In this way, the denoising UNet is trained both conditionally and unconditionally. Then at inference time, we simply use a guidance scale $s_g \ge 1$ to adjust the strength of conditional control over the predicted noise:
\begin{equation}
\label{eq:sg}
    \hat{\epsilon}_{\theta}(\mathbf{z}_t,\omega_{\theta'}(\mathcal{E}(\mathbf{g})))=\epsilon_{\theta}(\mathbf{z}_t,\varnothing)+s_\mathbf{g}\cdot(\epsilon_{\theta}(\mathbf{z}_t,\omega_{\theta'}(\mathcal{E}(\mathbf{g})))-\epsilon_{\theta}(\mathbf{z}_t,\varnothing)).
\end{equation}
In practice, we empirically set the outfitting dropout ratio to $10\%$ in training, and the guidance scale $s_\mathbf{g}$ to $1.2$ as default. We exclusively employ Outfitting Dropout within the Hydra Fusion Block that bridges HydraNet and MainNet. We have observed that applying outfitting dropout elsewhere introduces pixel-level artifacts.

\begin{table*}[t]
\centering
\caption{Full quantitative comparisons on the proprietary dataset. AnyFit \emph{(xxx)} represents AnyFit trained on the corresponding \emph{xxx} dataset.}
\setlength{\tabcolsep}{2.6mm}
  \begin{tabular}{l c c c c c c }
    \toprule
    \multicolumn{2}{c}{Method}                              
    & &LPIPS $\downarrow$ & SSIM $\uparrow$ & FID $\downarrow$ & KID $\downarrow$ \\
      \cmidrule{1-2} \cmidrule{4-7} 
      \multicolumn{2}{c}{LADI-VTON~\cite{morelli2023ladi}} 
    && 0.252 & 0.734 & 52.24 & 6.51 \\
    \multicolumn{2}{c}{DCI-VTON~\cite{gou2023dci}} 
    && 0.264 & 0.734 & 57.96 & 12.35  \\
    \multicolumn{2}{c}{StableVTON-base~\cite{kim2023stableviton}} 
    && 0.245 & 0.694 & 54.70 & 8.44 \\
    \multicolumn{2}{c}{StableVTON-repainting~\cite{kim2023stableviton}} 
    && 0.242 & 0.720 & 53.80 & 8.13 \\
    \multicolumn{2}{c}{IDM~\cite{choi2024idm}} 
    && 0.247 & 0.701 & 48.76 & 4.35 \\
      \cmidrule{1-2} \cmidrule{4-7}
    \multicolumn{2}{c}{\textbf{AnyFit (VITON-HD)}} 
    && \underline{0.200} & \underline{0.740} & \underline{46.95} & \underline{2.73} \\
    \multicolumn{2}{c}{\textbf{AnyFit (proprietary)}} 
    && \textbf{0.181} & \textbf{0.743} & \textbf{43.97} & \textbf{0.69} \\  
    \bottomrule
  \end{tabular}
  \label{tab:wild_supp}
\end{table*}

\section{More experiment results}
\label{appendix:exp_more}

\subsection{Full comparison on the proprietary dataset}
\begin{figure}[!ht]
  \centering
\includegraphics[width=1.0\textwidth]{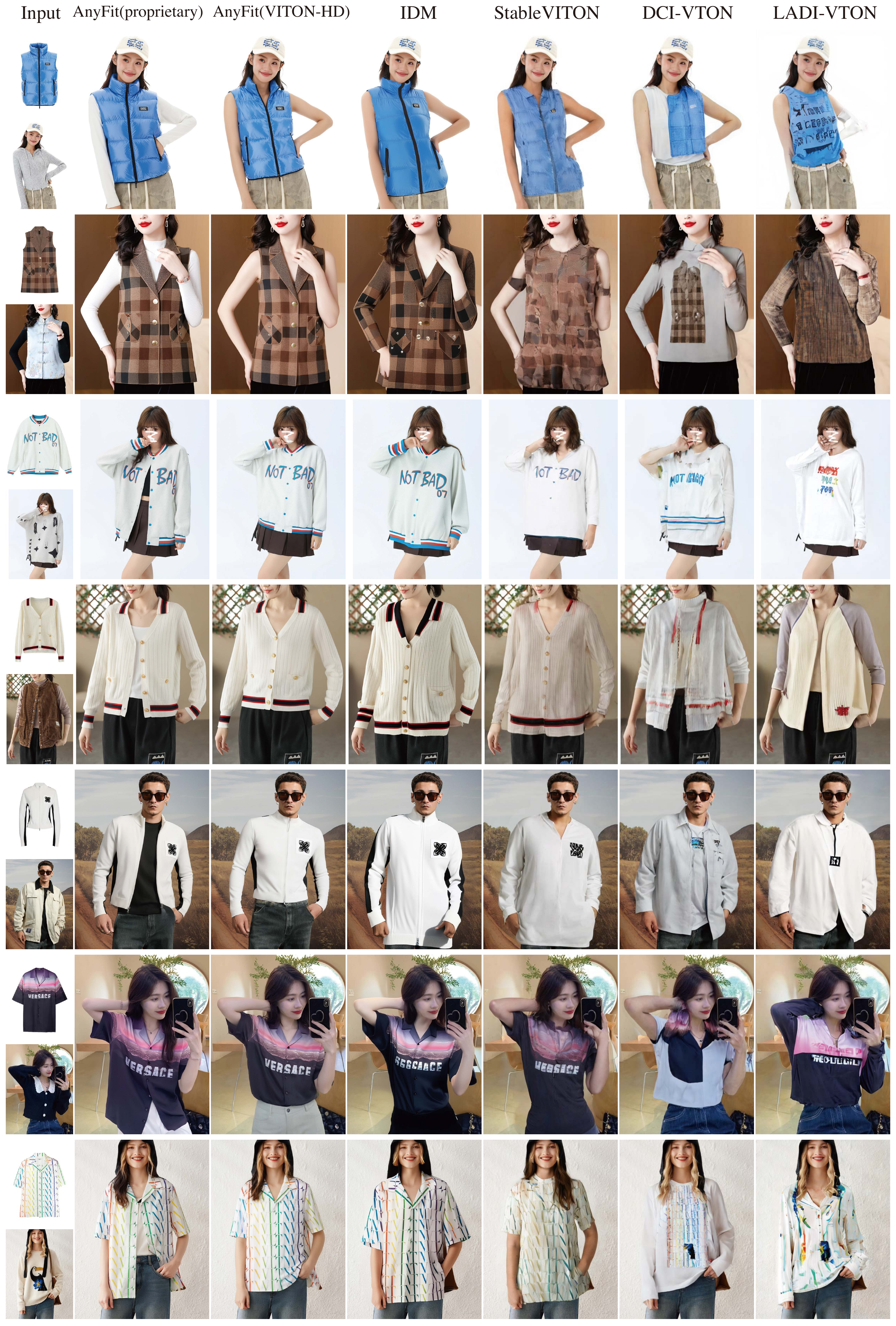}
  \caption{More visual comparisons on the proprietary dataset. AnyFit displays superior garment details and outfit styling. Best viewed when zoomed in.} 
  \label{fig:inthewild_comparison_supp}
\end{figure}

Due to space constraints, we have not presented quantitative comparisons on the proprietary dataset in a paired setting within the main body of the paper; they are additionally reported in Tab.~\ref{tab:wild_supp}. Furthermore, we showcase additional results in Fig.~\ref{fig:inthewild_comparison_supp}, which include outcomes from state-of-the-art baselines and AnyFit trained on both the proprietary dataset and VITON-HD. Our method exhibits a considerable lead in performance.

\subsection{More visual results}
\begin{figure}[!ht]
  \centering
\includegraphics[width=1.0\textwidth]{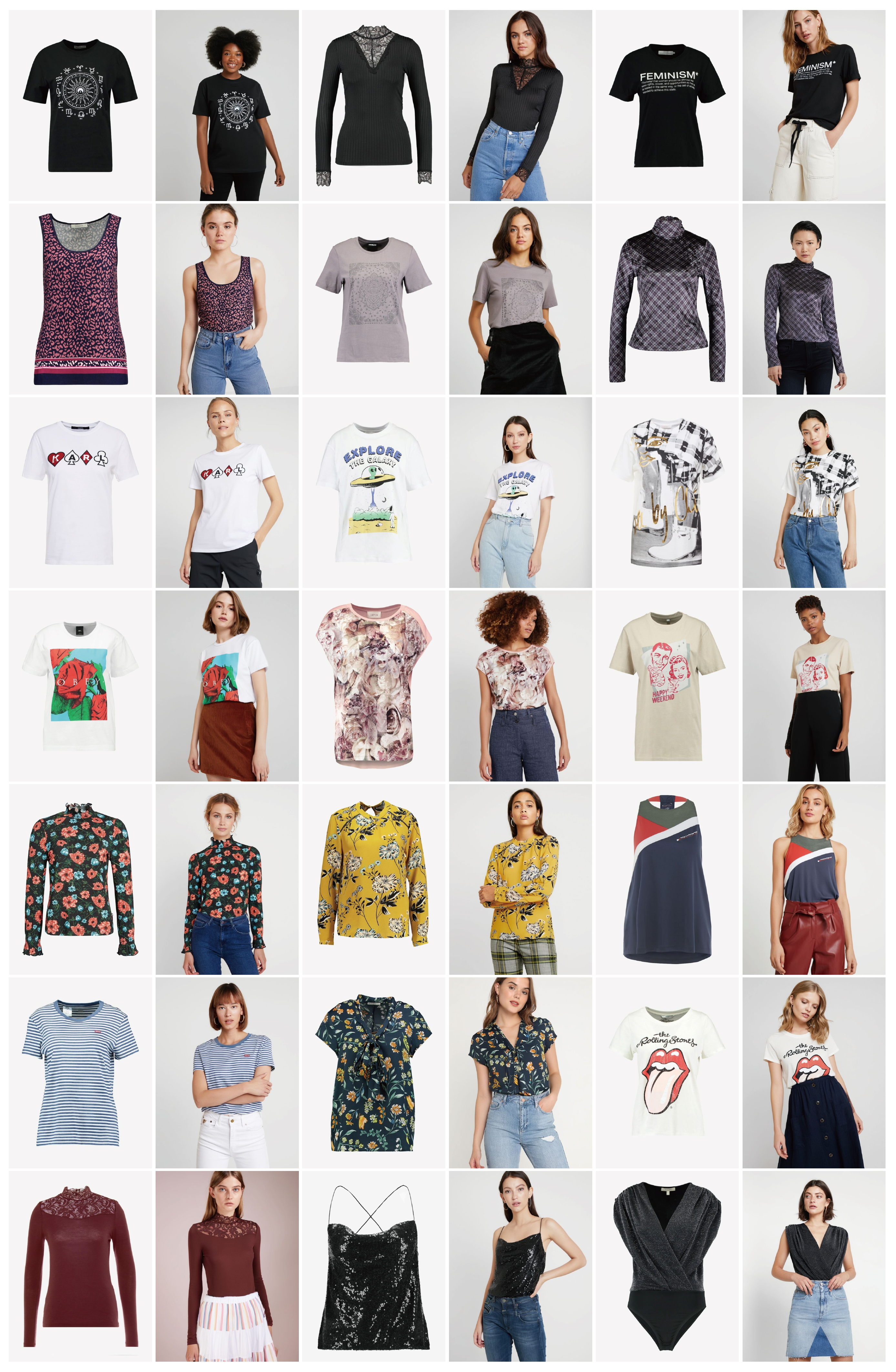}
  \caption{More visual results on the VITON-HD test data by AnyFit trained on VITON-HD training data. Best viewed when zoomed in.} 
  \label{fig:main_viton_supp}
\end{figure}

\begin{figure}[!ht]
  \centering
\includegraphics[width=1.0\textwidth]{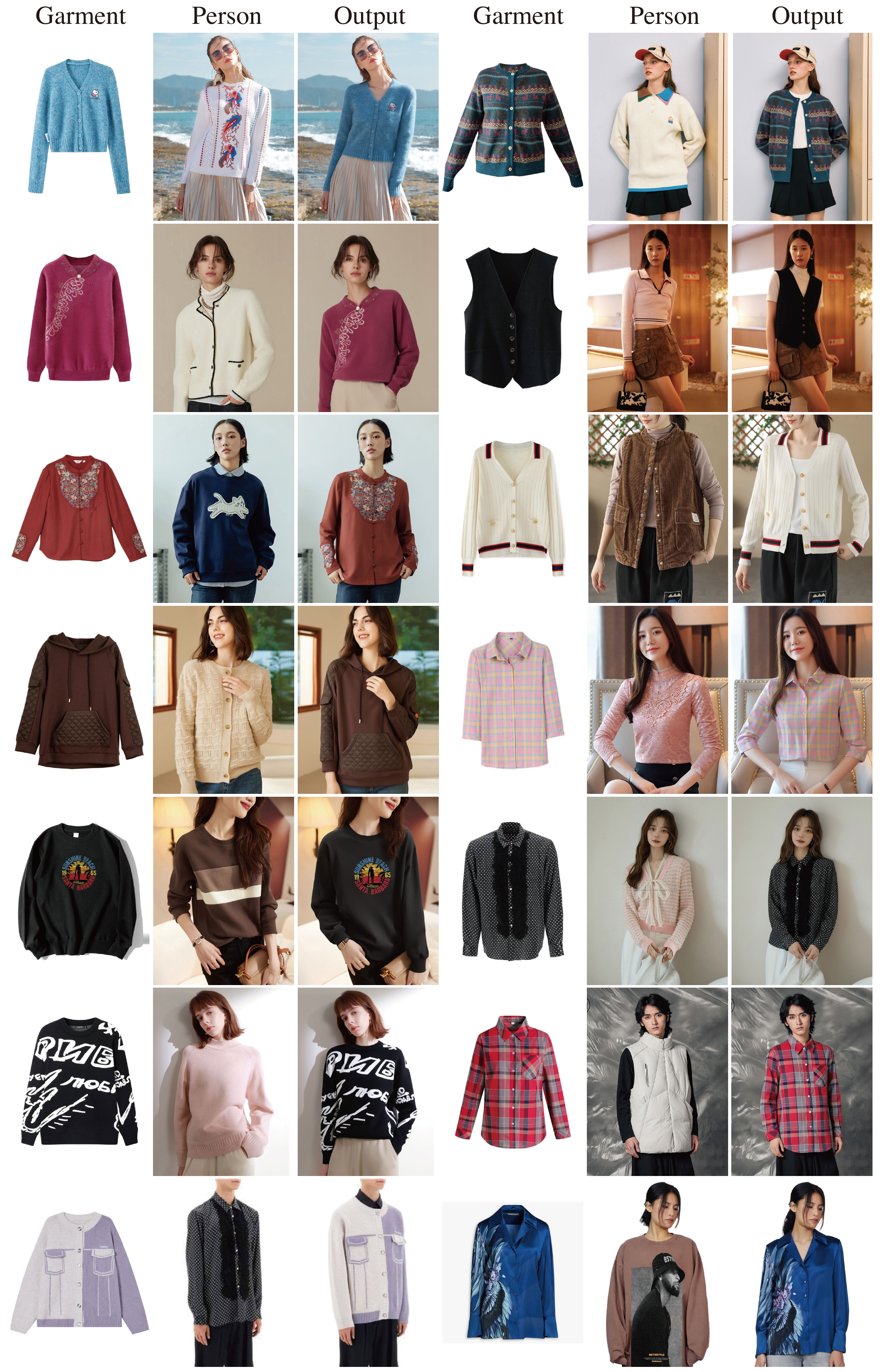}
  \caption{More visual results on the proprietary test data by AnyFit trained on proprietary training data. Best viewed when zoomed in.} 
  \label{fig:main_wild_supp}
\end{figure}

\begin{figure}[!ht]
  \centering
\includegraphics[width=1.0\textwidth]{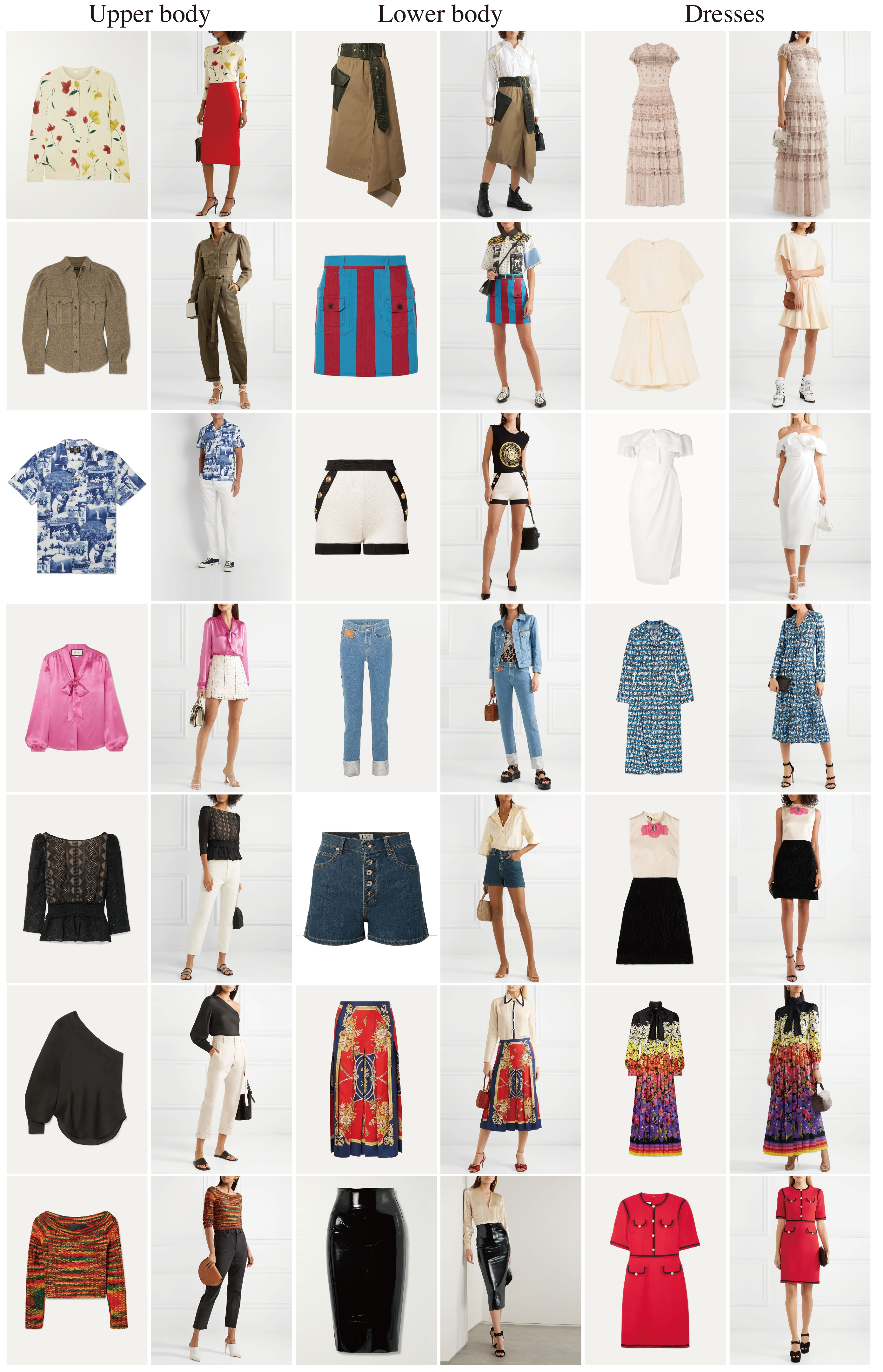}
  \caption{More visual results on the DressCode test data by AnyFit trained on DressCode training data. Best viewed when zoomed in.} 
  \label{fig:main_dc_supp}
\end{figure}

We provide more visual results on VITON-HD, DressCode, and proprietary dataset for inspection in Fig.~\ref{fig:main_viton_supp}, ~\ref{fig:main_wild_supp} and ~\ref{fig:main_dc_supp}. Fig.~\ref{fig:multi_ablation_supp} provides more results about multi-garment try-ons.

\clearpage

\end{document}